\documentclass[11pt,a4paper]{article}
\usepackage[hyperref]{acl2019}
\usepackage{times}
\usepackage{latexsym}
\usepackage{multirow}
\usepackage{float}
\usepackage{graphicx}
\usepackage{amsmath}
\usepackage{amsfonts}
\usepackage{amssymb}
\usepackage{todonotes}
\usepackage{xfrac}

\usepackage{graphicx}
\usepackage{subcaption}

\usepackage[utf8]{inputenc}
\usepackage[T1]{fontenc}
\usepackage[russian,english]{babel}
\usepackage{booktabs, tabularx}

\usepackage{url}

\aclfinalcopy % Uncomment this line for the final submission
\def\aclpaperid{1412} %  Enter the acl Paper ID here

\title{Analyzing Multi-Head Self-Attention:\\ Specialized Heads Do the Heavy Lifting, the Rest Can Be Pruned}

\author{Elena Voita$^{1,2}$ \quad David Talbot$^1$ \quad Fedor Moiseev$^{1,5}$ \quad Rico Sennrich$^{3,4}$ \quad Ivan Titov$^{3,2}$\bigskip\\
  $^1$Yandex, Russia \quad 
  $^2$University of Amsterdam, Netherlands \\% \smallskip\\
  $^3$University of Edinburgh, Scotland  \quad
  $^4$University of Zurich, Switzerland \\%\smallskip\\
  $^5$Moscow Institute of Physics and Technology, Russia\\
  {\tt \{lena-voita, talbot, femoiseev\}@yandex-team.ru}  \\ {\tt rico.sennrich@ed.ac.uk} \quad {\tt ititov@inf.ed.ac.uk}
	}
	
\date{}

\begin{document}
\maketitle
\begin{abstract}

Multi-head self-attention is a key component of the Transformer, a state-of-the-art architecture for neural machine translation. In this work we evaluate the contribution made by individual attention heads  in the encoder to the overall performance of the model and analyze the roles played by them. We find that the most important and confident heads play consistent and often linguistically-interpretable roles. When pruning heads using a method based on stochastic gates and a differentiable relaxation of the $L_0$ penalty, we observe that specialized heads are last to be pruned. Our novel pruning method removes the vast majority of heads without seriously affecting performance. 
For example, on the English-Russian WMT dataset, pruning 38 out of 48 encoder heads results in a drop of only 0.15 BLEU.\footnote{We release code at \url{https://github.com/lena-voita/the-story-of-heads}.}

\end{abstract}

\section{Introduction}

The Transformer~\cite{attention-is-all-you-need} has become the dominant modeling paradigm in neural machine translation. 
It follows the encoder-decoder framework using stacked multi-head self-attention and fully connected layers. Multi-head attention was shown to make more efficient use of the model's capacity: 
performance of the model with 8 heads is almost 1 BLEU point higher than that of a model of the same size with single-head attention~\cite{attention-is-all-you-need}.
The Transformer achieved state-of-the-art results in recent shared translation tasks \cite{bojar-EtAl:2018:WMT1,iwslt18-overview}. Despite the model's widespread adoption and recent attempts to investigate the kinds of information learned by the model's encoder~\cite{raganato-tiedemann:2018:BlackboxNLP}, 
the analysis of multi-head attention and its importance for translation is challenging. 
Previous analysis of multi-head attention considered the average of attention weights over all heads at a given position or focused only on the maximum attention weights~\cite{voita18,tang-sennrich-nivre:2018:WMT}, but neither method explicitly takes into account the varying importance of different heads.
Also, this obscures the roles played by individual heads which, as we show, influence the generated translations to differing extents.
We attempt to answer the following questions:
\begin{itemize}
    \item To what extent does translation quality depend on individual encoder heads?
    \item Do individual encoder heads play consistent and interpretable roles? If so, which are the most important ones for translation quality?
    \item Which types of model attention (encoder self-attention, decoder self-attention or decoder-encoder attention) are most sensitive to the number of attention heads and on which layers?
    \item Can we significantly reduce the number of attention heads while preserving translation quality?
\end{itemize}

We start by identifying the most important heads in each encoder layer 
using layer-wise relevance propagation~\cite{lrp-ding-2017}. 
For heads judged to be important, we then attempt to characterize the roles they perform. We observe the following types of role: positional (heads attending to an adjacent token), syntactic (heads attending to tokens in a specific syntactic dependency relation) and attention to rare words (heads pointing to the least frequent tokens in the sentence).

To understand whether the remaining heads perform vital but less easily defined roles, or are simply redundant to the performance of the model as measured by translation quality, we introduce a method for pruning heads based on~\citet{louizos2018learning}. While we cannot easily incorporate the number of active heads as a penalty term in our learning objective (i.e.\ the $L_0$ regularizer), we can use a differentiable relaxation. 
We prune attention heads in a continuous learning scenario starting from the converged full model 
and identify the roles of those which remain in the model. 
These experiments corroborate the findings of layer-wise relevance propagation;
in particular, heads with clearly identifiable positional and syntactic functions are pruned last and hence shown to be most important for the translation task.

Our key findings are as follows:
\begin{itemize}
    \item Only a small subset of heads are important for translation;
    \item Important heads have one or more specialized and interpretable functions in the model;
    \item The functions correspond to attention to neighbouring words and to tokens in specific syntactic dependency relations.
\end{itemize}

\section{Transformer Architecture}
In this section, we briefly describe the Transformer architecture~\cite{attention-is-all-you-need} introducing the terminology used in the rest of the paper.

The Transformer is an encoder-decoder model that uses stacked self-attention and fully connected layers for both the encoder and decoder.
The encoder consists of $N$ layers, each containing two sub-layers: (a) a multi-head self-attention mechanism, and (b) a feed-forward network.
The multi-head attention mechanism relies on scaled dot-product attention, which operates on a query $Q$, a key $K$ and a value $V$:
\begin{equation}
\textnormal{Attention}(Q, K, V ) = \textnormal{softmax}\left(\frac{QK^T}{\sqrt{d_k}}\right)V
\label{eq:mult_attention}
\end{equation}
where $d_k$ is the key dimensionality. In self-attention, queries, keys and values come from the output of the previous layer.

The multi-head attention mechanism obtains $h$ (i.e.\ one per head) different representations of ($Q$, $K$, $V$), computes scaled dot-product attention for each representation, concatenates the results, and projects the concatenation through a feed-forward layer. This can be expressed in the same notation as Equation~(\ref{eq:mult_attention}):
\begin{equation}
\textnormal{head}_i = \textnormal{Attention}(QW_i^Q , K W_i^K , V W_i^V )
\end{equation}
\vspace{-4ex}
\begin{equation}
\textnormal{MultiHead}(Q, K, V ) = \textnormal{Concat}_i(\textnormal{head}_i)W^O
\label{eq:concat_heads}
\end{equation}
where the $W_i$ and $W^O$ are parameter matrices. % that are learned. 

The second component of each layer of the Transformer network is a feed-forward network. The authors propose using a two-layer network with a ReLU activation. 

Analogously, each layer of the decoder contains the two sub-layers mentioned above as well as an additional multi-head attention sub-layer. This additional sub-layer receives the output of the encoder as its keys and values.

The Transformer uses multi-head attention in three different ways: encoder self-attention, decoder self-attention and decoder-encoder attention. In this work, we concentrate primarily on encoder self-attention.

\section{Data and setting}
\label{sect:data_setting}
We focus on English as a source language and consider three target languages: Russian, German and French. For each language pair, we use the same number of sentence pairs from WMT data to control for the amount of training data and train Transformer models with the same numbers of parameters. We use 2{.}5m sentence pairs, corresponding to the amount of English--Russian parallel training data (excluding UN and Paracrawl). In Section~\ref{sect:dependency_heads} we use the same held-out data for all language pairs; these are 50k English sentences taken from the WMT EN-FR data not used in training.

For English-Russian, we perform additional experiments using the publicly available OpenSubtitles2018 corpus~\cite{LISON18.294} to evaluate the impact of domains on our results. 

In Section~\ref{sect:pruning_attention_heads} we concentrate on English-Russian and two domains: WMT and OpenSubtitles.

Model hyperparameters, preprocessing and training details are provided in appendix~\ref{app:exp}.

\begin{figure*}[t!h!]
    \centering
    \begin{subfigure}[b]{0.3\textwidth}
        \includegraphics[width=\textwidth]{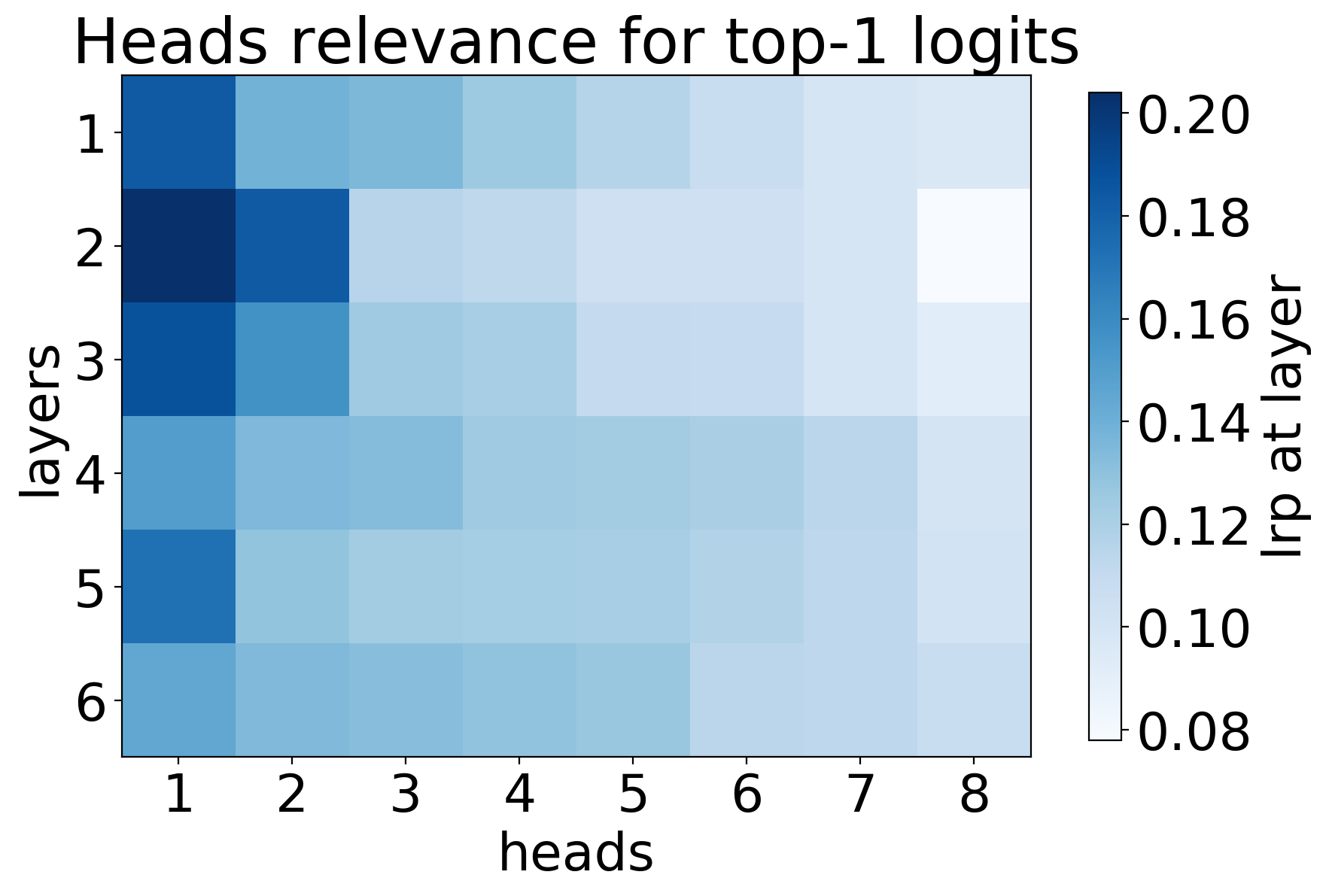}
        \caption{LRP}
        \label{fig:heads_lrp}
    \end{subfigure}
    \begin{subfigure}[b]{0.3\textwidth}
        \includegraphics[width=\textwidth]{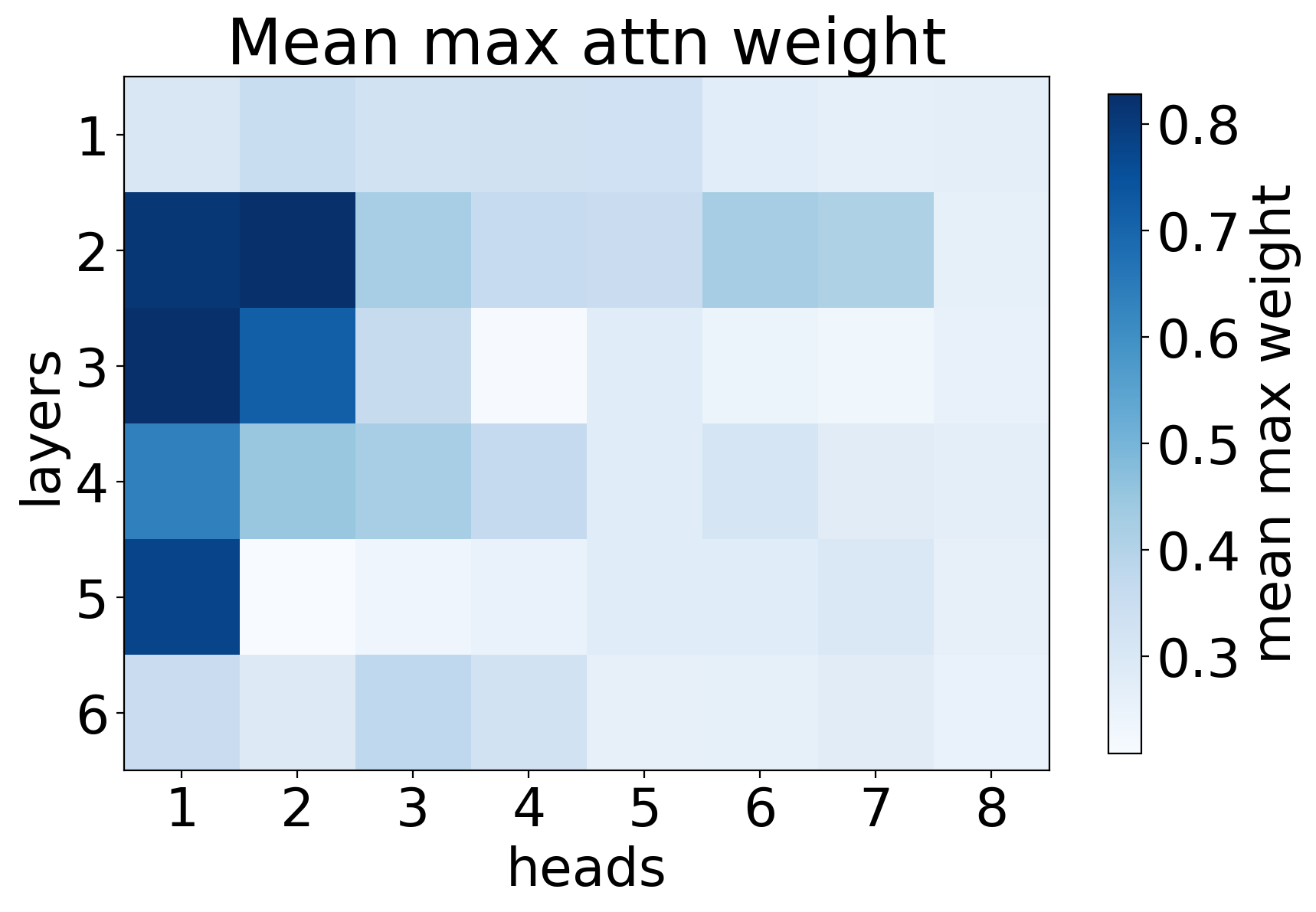}
        \caption{confidence}
        \label{fig:heads_mean_attn_max}
    \end{subfigure}
    \begin{subfigure}[b]{0.30\textwidth}
        \includegraphics[width=\textwidth]{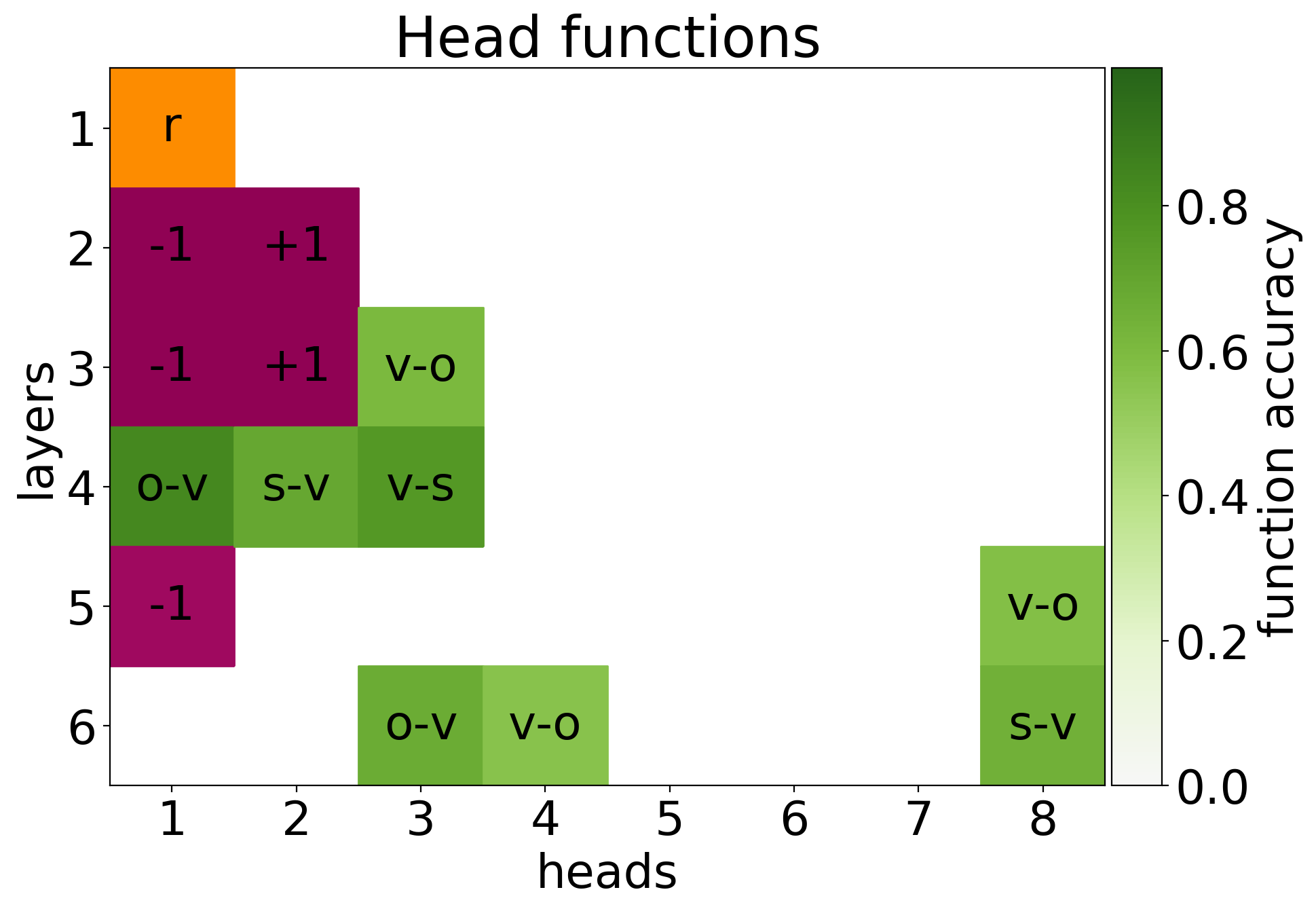}
        \caption{head functions}
        \label{fig:heads_functions}
    \end{subfigure}
    \caption{Importance (according to LRP), confidence, and function of self-attention heads. In each layer, heads are sorted by their relevance according to LRP. Model trained on 6m OpenSubtitles EN-RU data.}
    %\vspace{-2ex}
    \label{fig:heads_all_info}
\end{figure*}

\begin{figure}
    \centering
    \begin{subfigure}[b]{0.22\textwidth}
        %\centering
        \includegraphics[width=\textwidth]{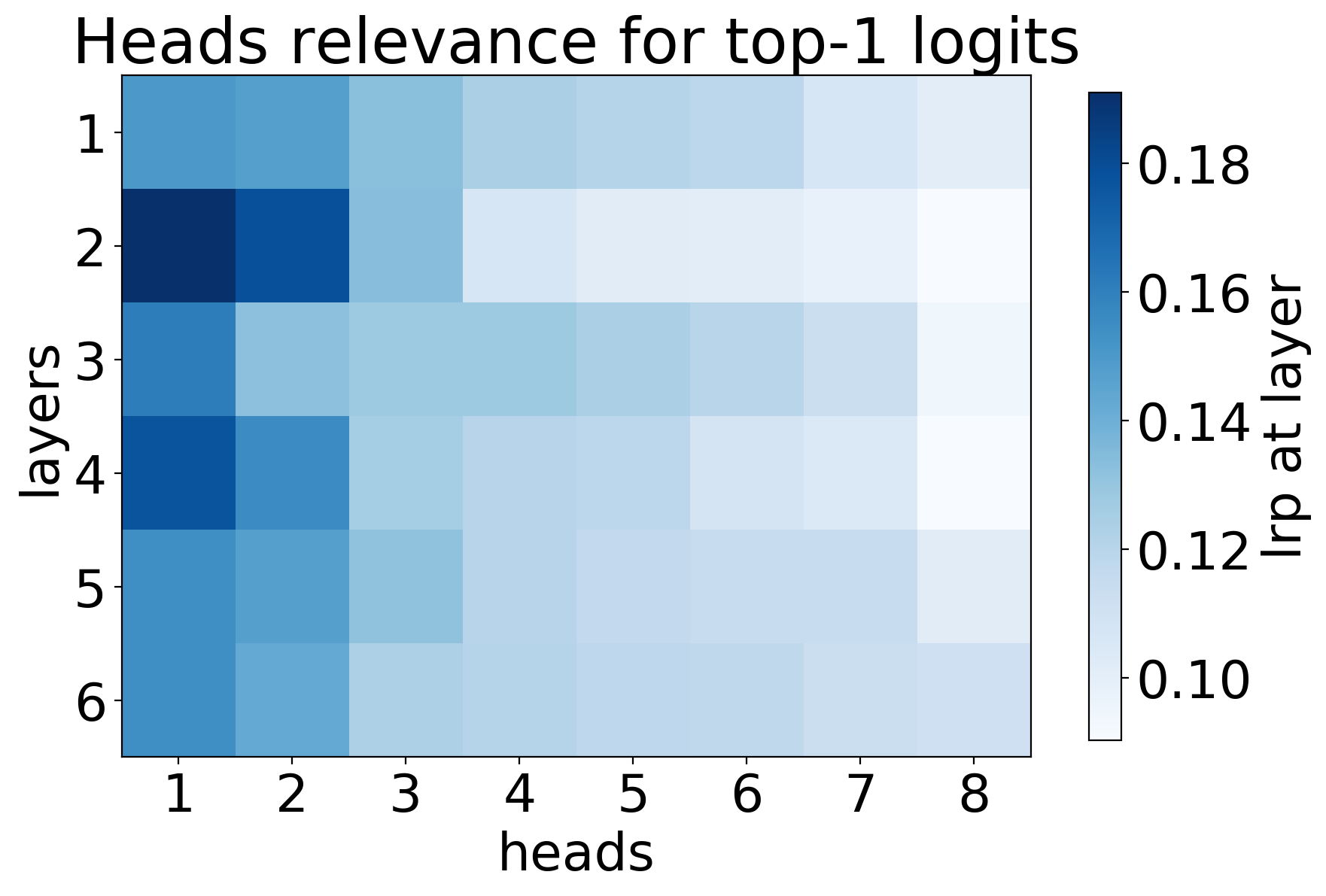}
        \caption{LRP (EN-DE)}
        \label{fig:heads_lrp_de}
    \end{subfigure}
    \begin{subfigure}[b]{0.22\textwidth}     \includegraphics[width=\textwidth]{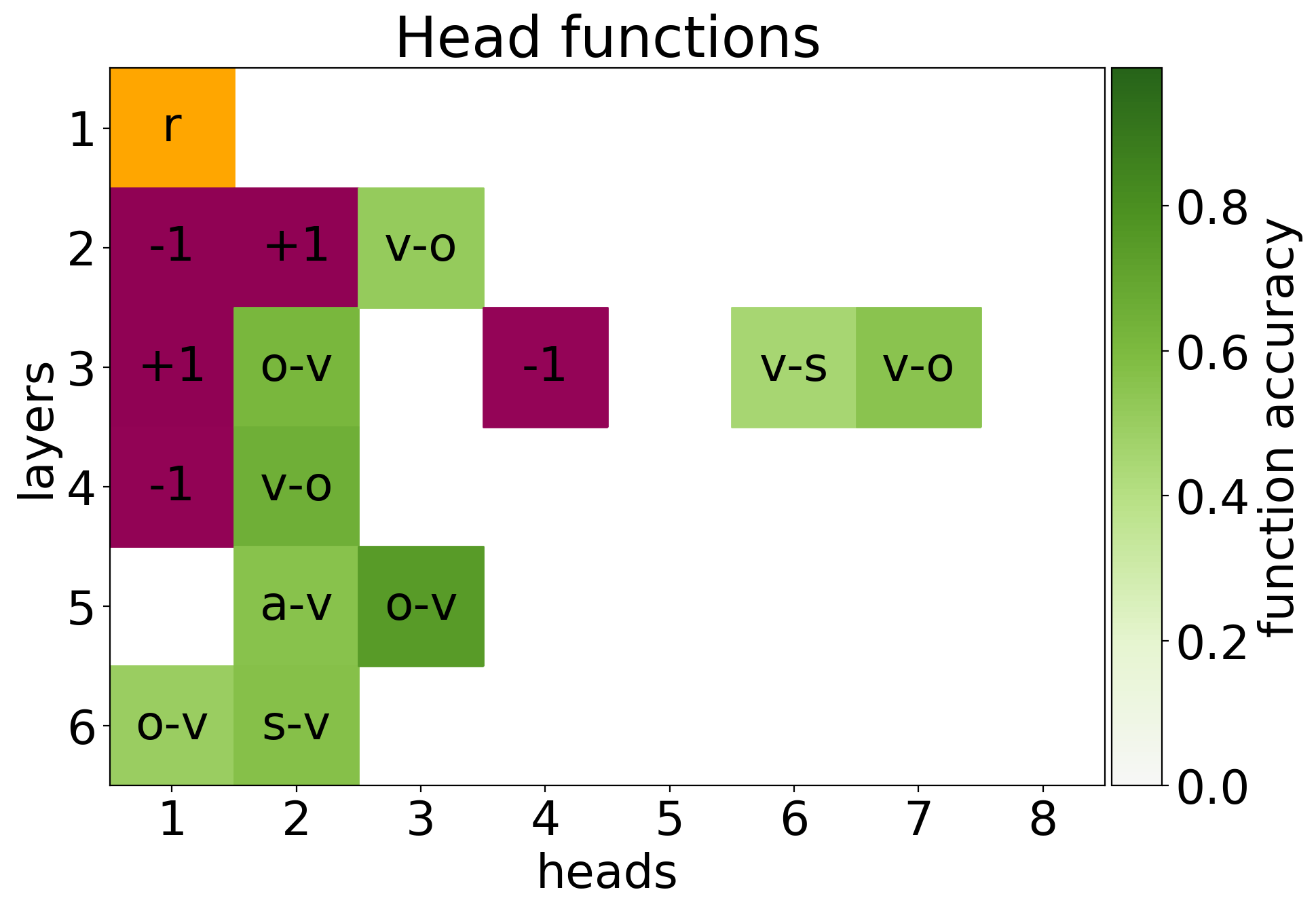}
        \caption{head functions}
        \label{fig:heads_functions_de}
    \end{subfigure}
    \vskip\baselineskip
    \begin{subfigure}[b]{0.22\textwidth} 
        \includegraphics[width=\textwidth]{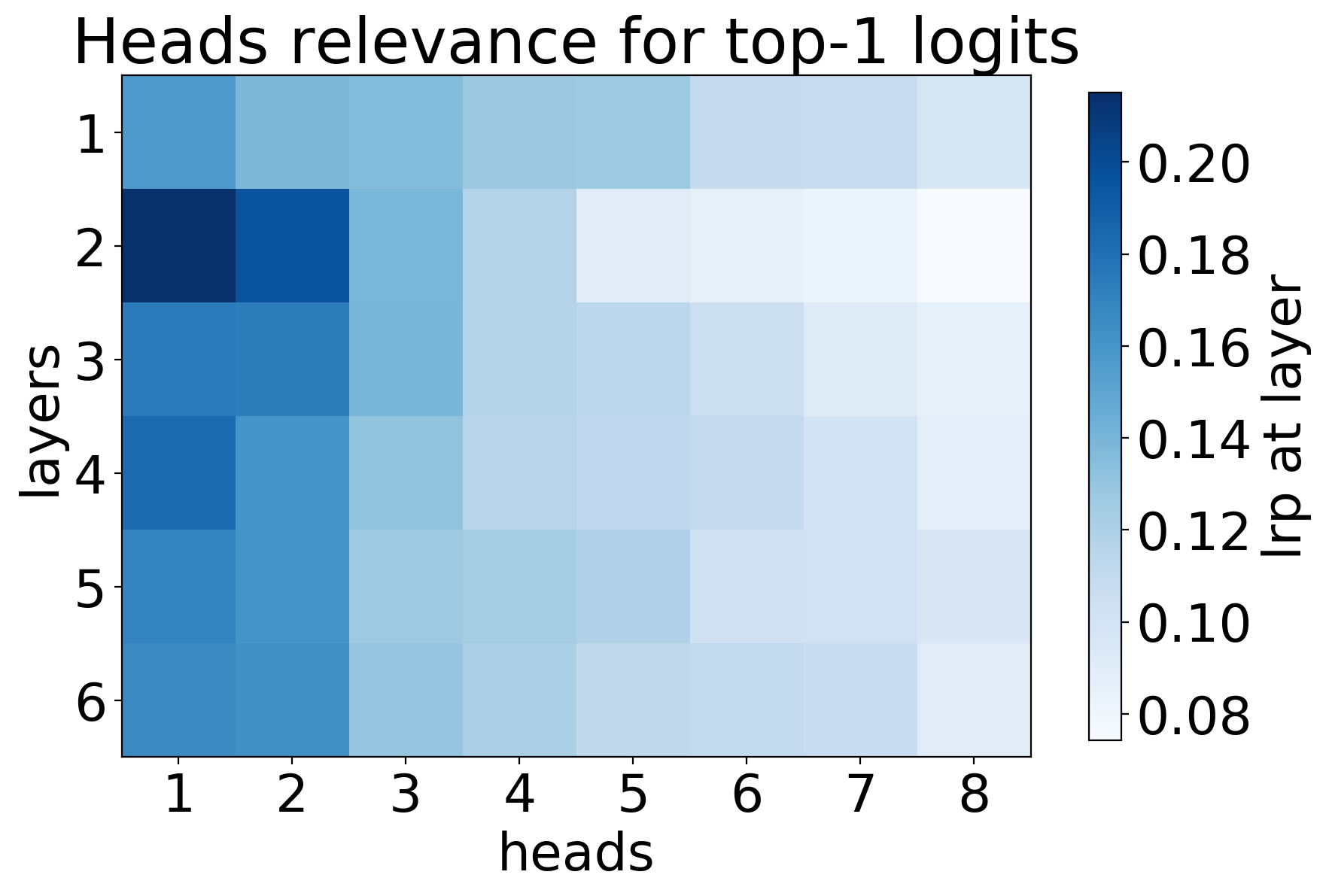}
        \caption{LRP (EN-FR)}
        \label{fig:heads_lrp_fr}
    \end{subfigure}
    \begin{subfigure}[b]{0.22\textwidth}
        \includegraphics[width=\textwidth]{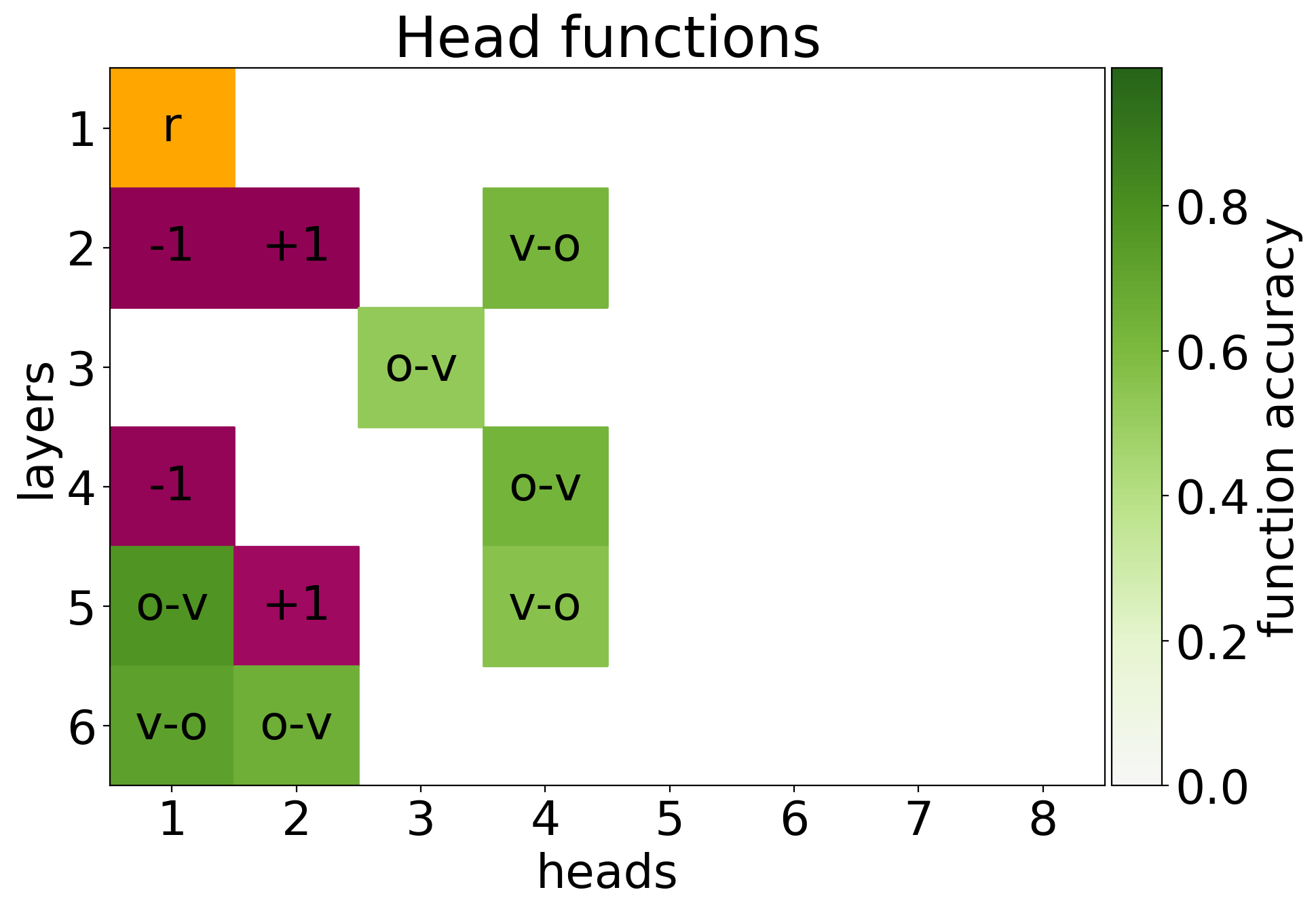}
        \caption{head functions}
        \label{fig:heads_functions_fr}
    \end{subfigure}
    \caption{Importance (according to LRP) and function of self-attention heads. In each layer, heads are sorted by their relevance according to LRP. Models trained on 2{.}5m WMT EN-DE (a, b) and EN-FR (c, d).} 
    %\vspace{-2ex}
    \label{fig:heads_all_info_de_fr}
\end{figure}

\section{Identifying Important Heads}
Previous work analyzing how representations are formed by the Transformer's multi-head attention mechanism focused on either the average or the maximum attention weights over all heads~\cite{voita18,tang-sennrich-nivre:2018:WMT}, but neither method explicitly takes into account the varying importance of different heads.
Also, this obscures the roles played by individual heads which, as we will show, influence the generated translations to differing extents.

We define the ``confidence'' of a head as the average of its maximum attention weight excluding the end of sentence symbol,\footnote{We exclude EOS on the grounds that it is not a real token.} where average is taken over tokens in a set of sentences used for evaluation (development set).
A confident head is one that usually assigns a high proportion of its attention to a single token. Intuitively, we might expect confident heads to be important to the translation task.  

Layer-wise relevance propagation (LRP)~\cite{lrp-ding-2017} is a method for computing the relative contribution of neurons at one point in a network to neurons at another.\footnote{A detailed description of LRP is provided in appendix \ref{app:lrp}.} 
Here we propose to use LRP to evaluate the degree to which different heads at each layer contribute to the top-1 logit predicted by the model. Heads whose outputs have a higher relevance value may be judged to be more important to the model's predictions.

The results of LRP are shown in Figures~\ref{fig:heads_lrp}, \ref{fig:heads_lrp_de}, \ref{fig:heads_lrp_fr}. 
In each layer, LRP ranks a small number of heads as much more important than all others.

The confidence for each head is shown in Figure~\ref{fig:heads_mean_attn_max}. 
We can observe that the relevance of a head as computed by LRP agrees to a reasonable extent with its confidence. The only clear exception to this pattern is the head judged by LRP to be the most important in the first layer. It is the most relevant head in the first layer but its average maximum attention weight is low. We will discuss this head further in Section~\ref{sec:topic_head}.

\section{Characterizing heads}
\label{sect:characterizing_heads}

We now turn to investigating whether heads play consistent and interpretable roles within the model.

We examined some attention matrices paying particular attention to heads ranked highly by LRP and identified three functions which heads might be playing:
\begin{enumerate}
\item positional: the head points to an adjacent token,
\item syntactic: the head points to tokens in a specific syntactic relation,
\item rare words: the head points to the least frequent tokens in a sentence. 
\end{enumerate}

Now we discuss the criteria used to determine if a head is performing one of these functions and examine properties of the corresponding heads.

\subsection{Positional heads}\label{subsec:positional_heads}
We refer to a head as ``positional'' if at least 90\% of the time its maximum attention weight is assigned to a specific relative position (in practice either -1 or +1, i.e.\ attention to adjacent tokens). Such heads are shown in purple in Figures~\ref{fig:heads_functions} for English-Russian, \ref{fig:heads_functions_de} for English-German, \ref{fig:heads_functions_fr} for English-French and marked with the relative position.

As can be seen, the positional heads correspond to a large extent to the most confident heads and the most important heads as ranked by LRP. In fact, the average maximum attention weight exceeds $0{.}8$ for every positional head for all language pairs considered here.

\subsection{Syntactic heads}
\label{sect:dependency_heads}

We hypothesize that, when used to perform translation, the Transformer's encoder may be responsible for disambiguating the syntactic structure of the source sentence. We therefore wish to know whether a head attends to tokens corresponding to any of the major syntactic relations in a sentence. 
In our analysis, we looked at the following dependency relations: 
nominal subject (nsubj), direct object (dobj), adjectival modifier (amod) and adverbial modifier (advmod). These include the main verbal arguments of a sentence and some other common relations. They also include those relations which might inform morphological agreement or government in one or more of the target languages considered here.

\subsubsection{Methodology}
We evaluate to what extent each head in the Transformer's encoder accounts for a specific dependency relation by comparing its attention weights to a predicted dependency structure generated using CoreNLP~\cite{manning-EtAl:2014:P14-5} 
on a large number of held-out sentences. 
We calculate for each head how often it assigns its maximum attention weight (excluding EOS) to a token with which it is in one of the aforementioned dependency relations. We count each relation separately and allow the relation to hold in either direction between the two tokens.

We refer to this relative frequency as the ``accuracy'' of head on a specific dependency relation in a specific direction.
Note that under this definition, we may evaluate the accuracy of a head for multiple dependency relations.

\begin{figure}[t!]
\center{\includegraphics[scale=0.18]{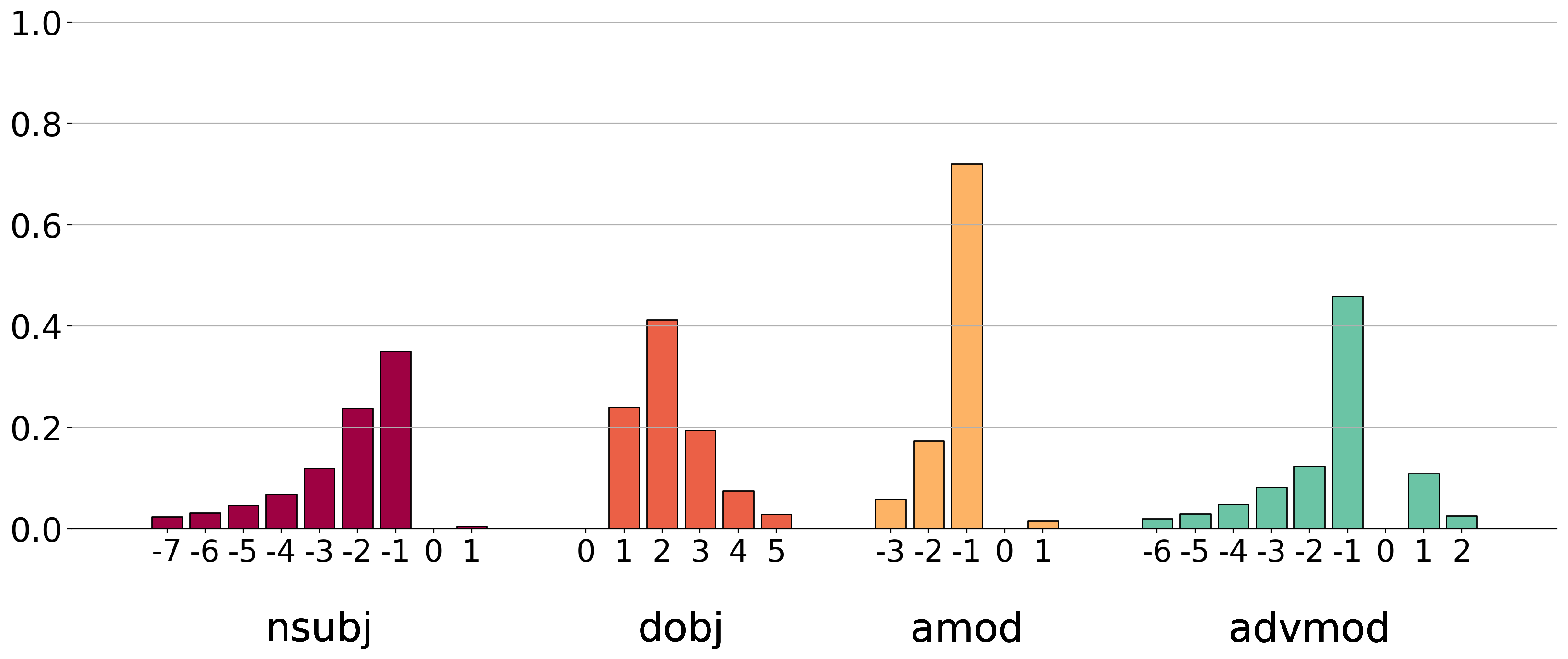}}
\caption{Distribution of the relative position of dependent for different dependency relations (WMT).}
\label{fig:dep_posdiff_distribution}
\end{figure}

Many dependency relations are frequently observed in specific relative positions (for example, often they hold between adjacent tokens, see Figure~\ref{fig:dep_posdiff_distribution}). We say that a head is ``syntactic'' if its accuracy is at least $10\%$ higher than the baseline that looks at the most frequent relative position for this dependency relation.

\subsubsection{Results}
\label{sec:syntactic-heads:results}

\begin{table}[t!]
\centering
\begin{tabular}{lccc}
\toprule
\bf dep.\!\!&\!\!\!\! \bf direction & \multicolumn{2}{c}{\bf best head / baseline}\\ 
 &  & \multicolumn{2}{c}{\bf accuracy} \\ 
 \cmidrule{1-4}
 & & \bf WMT  & \bf OpenSubtitles\!\!\!\\
\toprule
\multicolumn{3}{l}{nsubj}\\
&\!\!\!\!v $\rightarrow$ s & 45 / 35 & 77 / 45 \\
&\!\!\!\!s $\rightarrow$ v & 52 / 35 & 70 / 45\\
\cmidrule{1-4}

\multicolumn{3}{l}{dobj}\\
&\!\!\!\!v $\rightarrow$ o & 78 / 41 & 61 / 46\\
&\!\!\!\!o $\rightarrow$ v & 73 / 41 & 84 / 46\\
\cmidrule{1-4}

\multicolumn{3}{l}{amod}\\
   &\!\!\!\!noun $\rightarrow$ adj.m. & 74 / 72 & 81 / 80\\
&\!\!\!\!adj.m. $\rightarrow$ noun  & 82 / 72 & 81 / 80 \\
\cmidrule{1-4}

\multicolumn{3}{l}{advmod}\\
   &\!\!\!\!v $\rightarrow$ adv.m. & 48 / 46 & 38 / 33\\
&\!\!\!\!adv.m. $\rightarrow$ v   & 52 / 46 & 42 / 33\\

\bottomrule
\end{tabular}\textbf{}
\caption{Dependency scores for EN-RU, comparing the best self-attention head to a positional baseline. Models trained on 2{.}5m WMT data and 6m OpenSubtitles data.} \label{tab:dependency_head_scores}
\vspace{1ex}
\end{table}

\begin{figure}[t!]
\center{\includegraphics[scale=0.18]{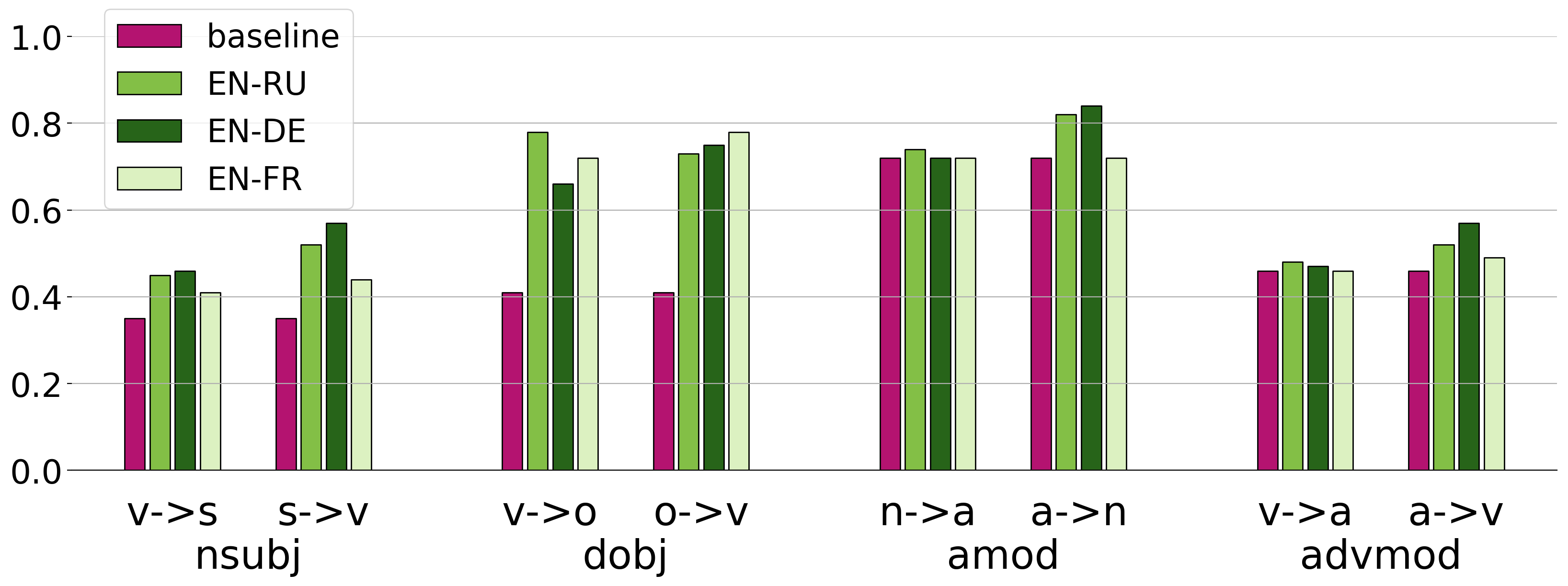}}
\caption{Dependency scores for EN-RU, EN-DE, EN-FR each trained on 2{.}5m WMT data. 
}
\label{fig:wmt_deps_ru_de_fr}
\end{figure}

\begin{figure*}[t!]
    \centering
    \begin{subfigure}[b]{0.30\textwidth}
        \includegraphics[width=\textwidth]{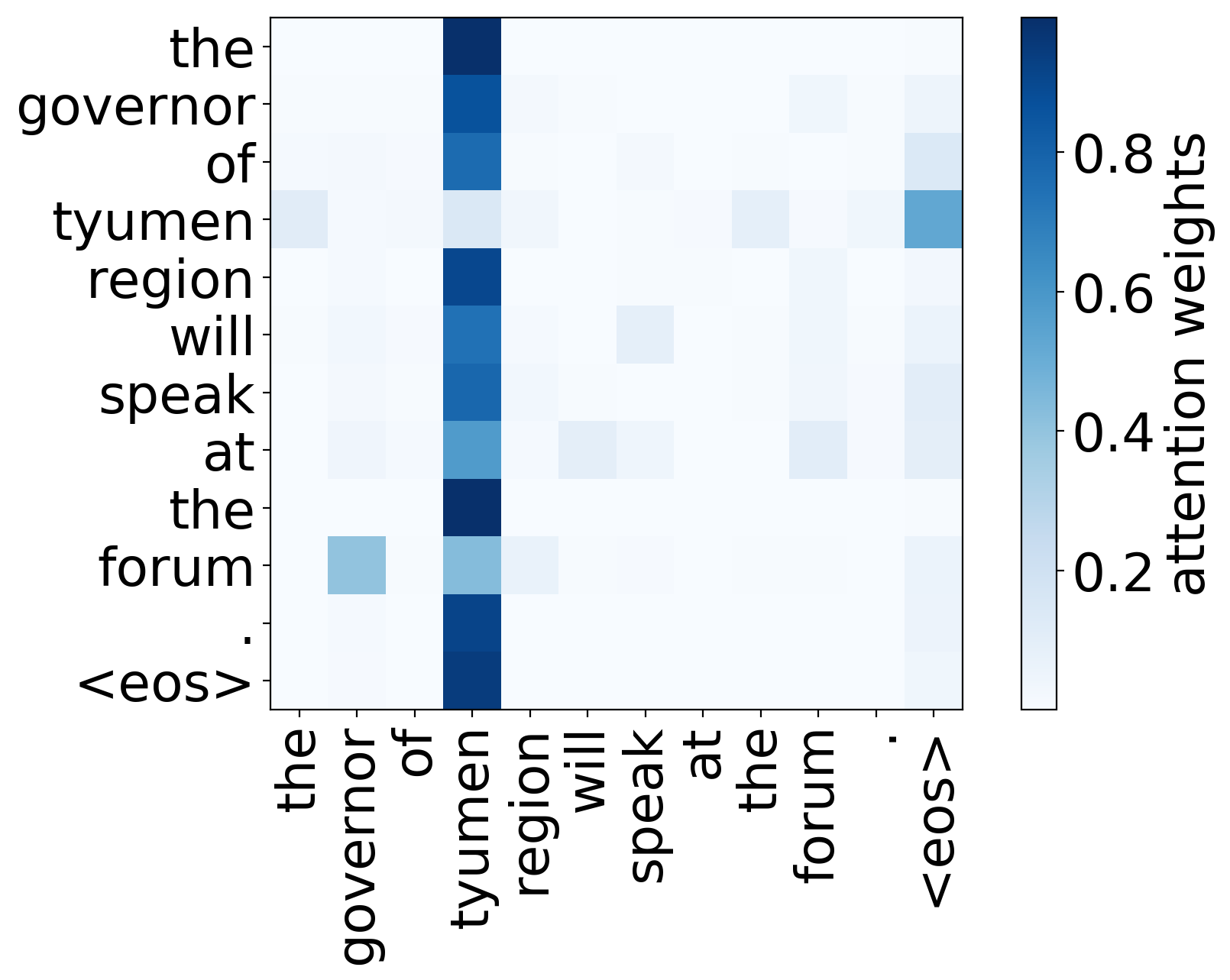}
        \caption{}
        \label{fig:topic_head_wmt_en_ru}
    \end{subfigure}
    \begin{subfigure}[b]{0.30\textwidth}
        \includegraphics[width=\textwidth]{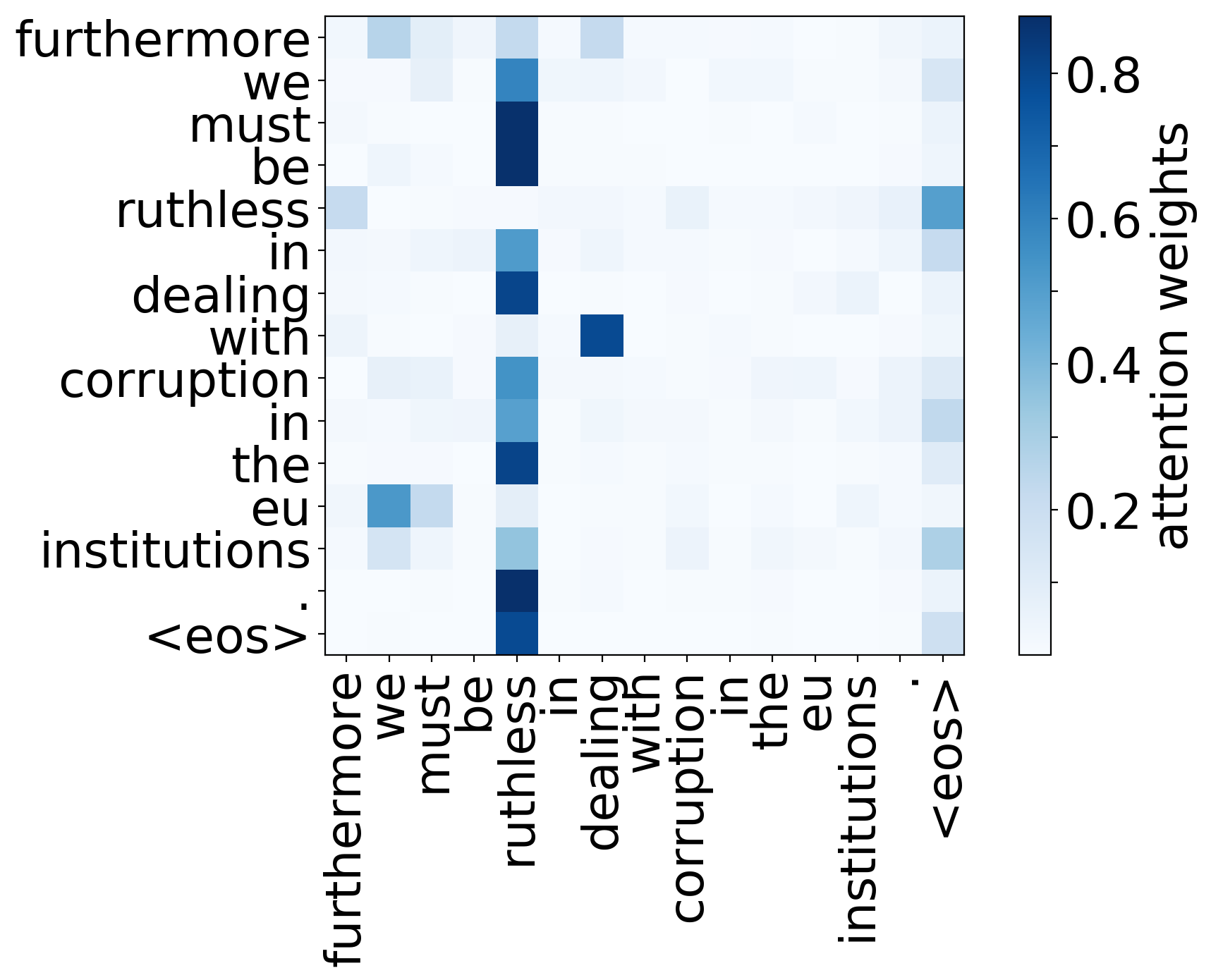}
        \caption{}
        \label{fig:topic_head_wmt_en_de}
    \end{subfigure}
    \begin{subfigure}[b]{0.30\textwidth}
        \includegraphics[width=\textwidth]{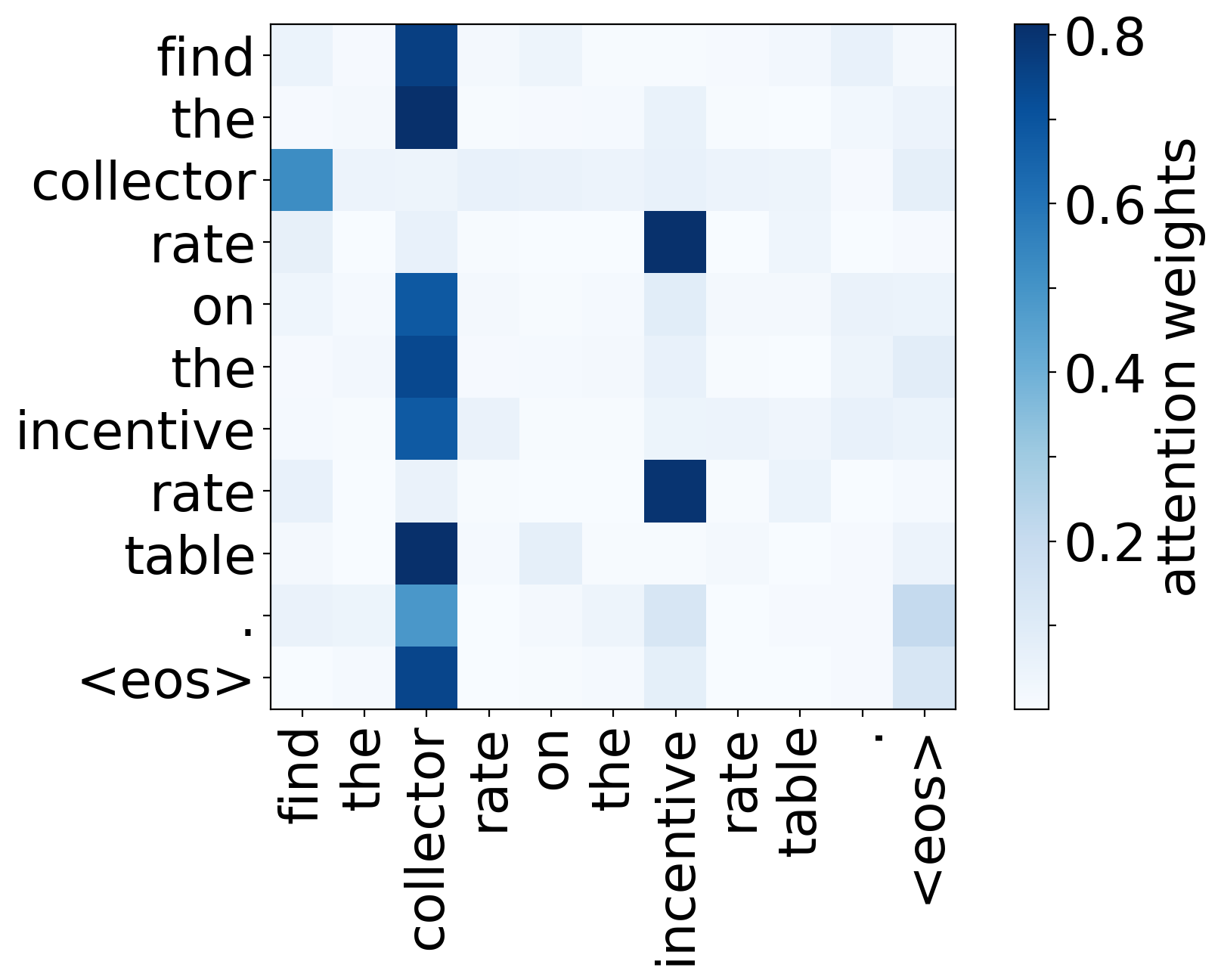}
        \caption{}
        \label{fig:topic_head_wmt_en_fr}
    \end{subfigure}
    \caption{Attention maps of the rare words head. Models trained on WMT: (a) EN-RU, (b) EN-DE, (c) EN-FR}\label{fig:topic_head}
\end{figure*}

Table~\ref{tab:dependency_head_scores} shows the accuracy of the most accurate head for each of the considered dependency relations on the two domains for English-Russian. Figure~\ref{fig:wmt_deps_ru_de_fr} compares the scores of the models trained on WMT with different target languages. 

Clearly certain heads learn to detect syntactic relations with accuracies significantly higher than the positional baseline. This supports the hypothesis that the encoder does indeed perform some amount of syntactic disambiguation of the source sentence. 

Several heads appear to be responsible for the same dependency relation. These heads are shown in green in Figures~\ref{fig:heads_functions}, \ref{fig:heads_functions_de}, \ref{fig:heads_functions_fr}.

Unfortunately, it is not possible to draw any strong conclusions from these results regarding the impact of target language morphology on the accuracy of the syntactic attention heads although relations with strong target morphology are among those that are most accurately learned. 

Note the difference in accuracy of the verb-subject relation heads across the two domains for English-Russian. We hypothesize that this is due to the greater variety of grammatical person present\footnote{First, second and third person subjects are encountered in approximately $6\%$, $3\%$ and $91\%$ of cases in WMT data and in $32\%$, $21\%$ and $47\%$ of cases in OpenSubtitles data.} in the Subtitles data which requires more attention to this relation. However, we leave proper analysis of this to future work.

\subsection{Rare words}
\label{sec:topic_head}

In all models (EN-RU, EN-DE, EN-FR on WMT and EN-RU on OpenSubtitles), we find that one head in the first layer is judged to be much more important to the model's predictions 
than any other heads in this layer.  

We find that this head points to the least frequent tokens in a sentence. For models trained on OpenSubtitles, among sentences where the least frequent token in a sentence is not in the top-500 most frequent tokens, this head points to the rarest token in 66$\%$ of cases, and to one of the two least frequent tokens in 83$\%$ of cases. For models trained on WMT, this head points to one of the two least frequent tokens in more than 50$\%$ of such cases. This head is shown in orange in Figures~\ref{fig:heads_functions}, \ref{fig:heads_functions_de}, \ref{fig:heads_functions_fr}. Examples of attention maps for this head for models trained on WMT data with different target languages are shown in Figure~\ref{fig:topic_head}.

\section{Pruning Attention Heads}
\label{sect:pruning_attention_heads}

We have identified certain functions of the most relevant heads at each layer and showed that to a large extent they are interpretable. What of the remaining heads? Are they redundant to translation quality or do they play equally vital but simply less easily defined roles? We introduce a method for pruning attention heads to try to answer these questions. Our method is based on \citet{louizos2018learning}. Whereas they pruned individual neural network weights, we prune entire model components (i.e.\ heads).  We start by describing our method and then examine how performance changes as we remove heads, identifying the functions of heads retained in the sparsified models.

\subsection{Method}

We modify the original Transformer architecture by multiplying the representation computed by each head$_i$ by a scalar gate $g_i$. Equation~(\ref{eq:concat_heads}) turns into\vspace{-2ex}
\begin{equation}
\nonumber
\textnormal{MultiHead}(Q,K,V )\!=\! \textnormal{Concat}_i(g_i\!\cdot\!\textnormal{head}_i)W^O.
\end{equation}
Unlike usual gates, $g_i$ are parameters specific to heads and are independent of the input (i.e.\ the sentence).
As we would like to disable less important heads completely rather than simply downweighting them,  we would ideally apply $L_0$ regularization to the scalars $g_i$. The $L_0$ norm equals the number of non-zero components and would push the model to switch off less important heads:
\vspace{-1ex}
$$
L_0(g_1, \ldots, g_h) = \sum_{i=1}^{h} (1 - [[ g_i = 0 ]]),
$$
where $h$ is the number of heads, and $[[ \quad ]]$ denotes the indicator function. 

\begin{figure}[t!]
    \centering
    \begin{subfigure}[b]{0.23\textwidth}
        \includegraphics[width=\textwidth]{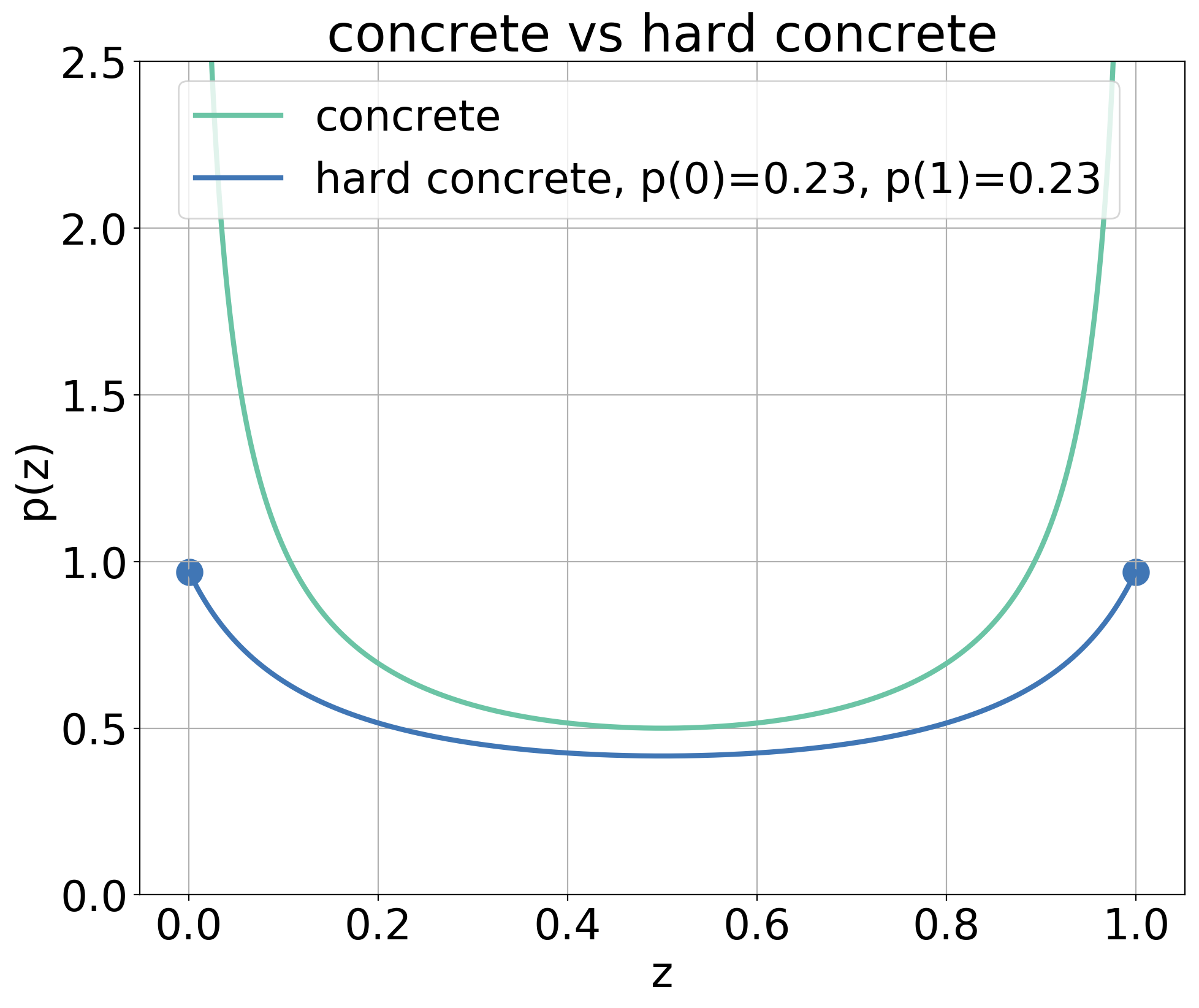}
        \caption{}
        \label{fig:hard_concrete}
    \end{subfigure}
    \begin{subfigure}[b]{0.23\textwidth}
       \includegraphics[width=\textwidth]{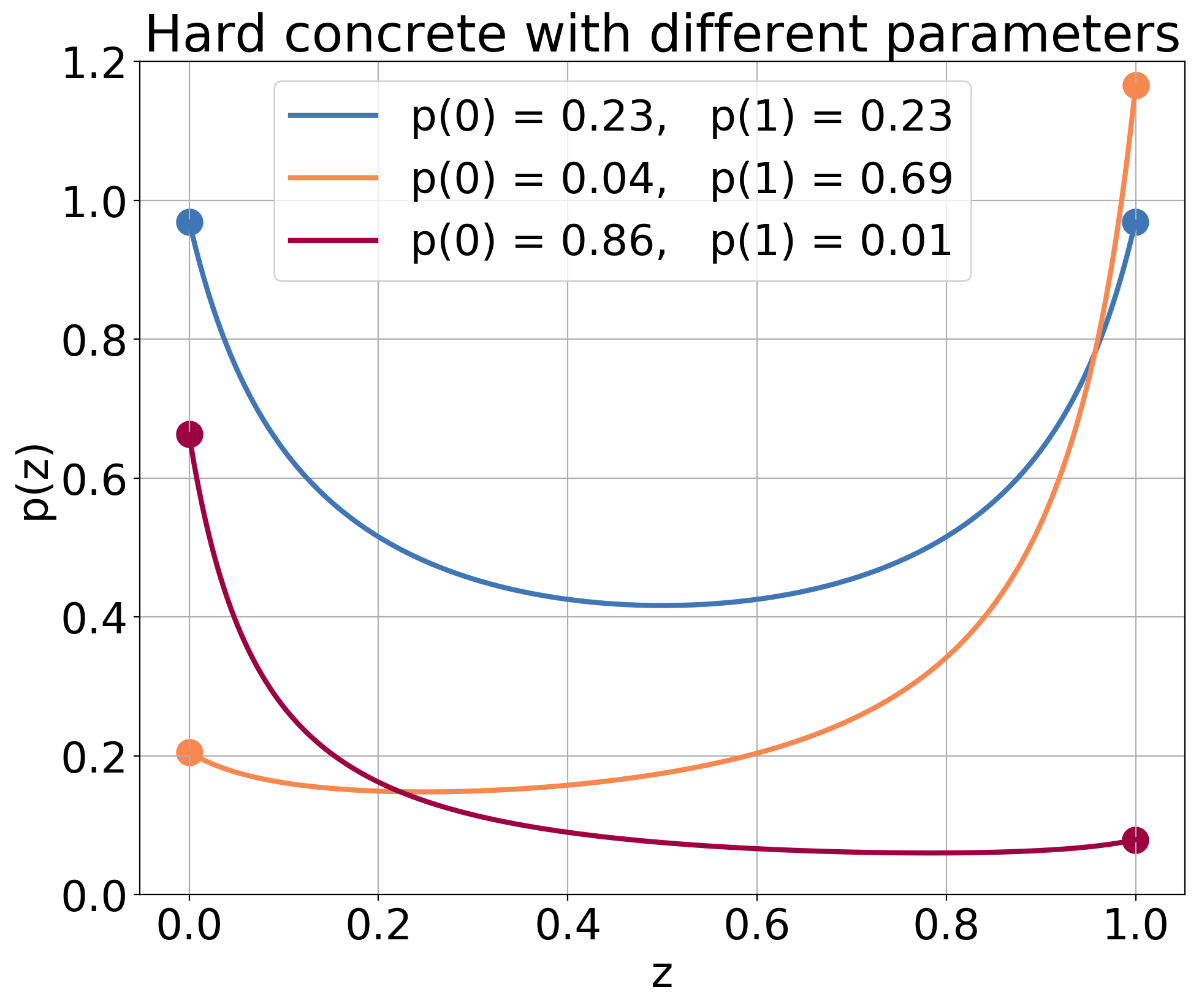}
        \caption{}
        \label{fig:hard_concrete_params}
    \end{subfigure}
    \caption{Concrete distribution: (a) Concrete and its stretched and rectified version (Hard Concrete); (b) Hard Concrete distributions with different parameters.}
    \label{fig:concrete}
\end{figure}

Unfortunately, the $L_0$ norm is non-differentiable and so cannot be directly incorporated as a regularization term in the objective function. 
Instead, we use a stochastic relaxation: each gate $g_i$ is now a  random variable drawn independently from a head-specific distribution.\footnote{In training, we resample gate values $g_i$ for each batch.} 
We use the Hard Concrete distributions~\cite{louizos2018learning},  a parameterized family of mixed discrete-continuous distributions over the closed interval $[0,1]$, see Figure~\ref{fig:hard_concrete}. The distributions have non-zero probability mass at 0 and 1, $P(g_i = 0 | \phi_i)$ and $P(g_i = 1 | \phi_i)$, where $\phi_i$ are the distribution parameters.  Intuitively, the Hard Concrete distribution is obtained by stretching the binary version of the Concrete (aka Gumbel softmax) distribution \cite{maddison2017concrete,jang2017gumbel} from the original support of $(0, 1)$ to $(- \epsilon, 1 + \epsilon)$ and then collapsing the probability mass assigned to $(- \epsilon, 1]$ and  $[1, 1 + \epsilon)$ to single points, 0 and 1, respectively. These stretching and rectification operations yield a mixed discrete-continuous distribution over $[0, 1]$.
Now the sum of the probabilities of heads being non-zero can be used as a relaxation of the $L_0$ norm:
\vspace{-1ex}
$$
\vspace{-1ex}
L_C(\phi) = \sum_{i=1}^{h} (1 - P(g_i = 0 | \phi_i)).
$$
The new training objective is
$$
\vspace{-1ex}
L(\theta, \phi) = L_{xent}(\theta, \phi) + \lambda L_C(\phi),
$$
where $\theta$ are the parameters of the original Transformer, $L_{xent}(\theta, \phi)$ is cross-entropy loss for the translation model, and $L_C(\phi)$ is the regularizer described above.  
The objective is easy to optimize: the reparameterization trick ~\cite{kingma-welling-vae,pmlr-v32-rezende14} can be used to backpropagate through the sampling process for each $g_i$, whereas the regularizer and its gradients are available in the closed form.  
Interestingly, we observe that the model converges to solutions where gates are either almost completely closed (i.e.\ the head is pruned, $P(g_i = 0 | \phi_i) \approx 1$) or completely open ($P(g_i = 1 | \phi_i) \approx 1$), the latter not being explicitly encouraged.\footnote{The `noise' pushes the network not to use middle values. The combination of noise and rectification has been previously used  to achieve discretization~(e.g., \citet{kaiser2018discrete}).}
This means that at test time we can treat the model as a standard Transformer and use only a subset of heads.\footnote{At test time, gate values are either 0 or 1 depending on which of the values $P(g_i = 0 | \phi_i)$, $P(g_i = 1 | \phi_i)$ is larger.}

When applying this regularizer, we start from the converged model trained without the $L_C$ penalty  (i.e.\ parameters~$\theta$ are initialized with the parameters of the converged model) and then add the gates and continue training the full objective.
By varying the coefficient $\lambda$ in the optimized objective, we obtain models with different numbers of heads retained.

\subsection{Pruning encoder heads}

To determine which head functions are most important in the encoder 
and how many heads the model needs, we conduct a series of experiments with gates applied only to encoder self-attention. Here we prune a model by fine-tuning a trained model with the regularized objective.\footnote{In preliminary experiments, we observed that fine-tuning a trained model gives slightly better results (0{.}2--0{.}6 BLEU) than applying the regularized objective, or training a model with the same number of self-attention heads, from scratch.} 
During pruning, the parameters of the decoder are fixed and only the encoder parameters and head gates are fine-tuned. By not fine-tuning the decoder, we ensure that the functions of the pruned encoder heads do not migrate to the decoder.

\subsubsection{Quantitative results: BLEU score}

\begin{figure}[t!]
\center{\includegraphics[scale=0.28]{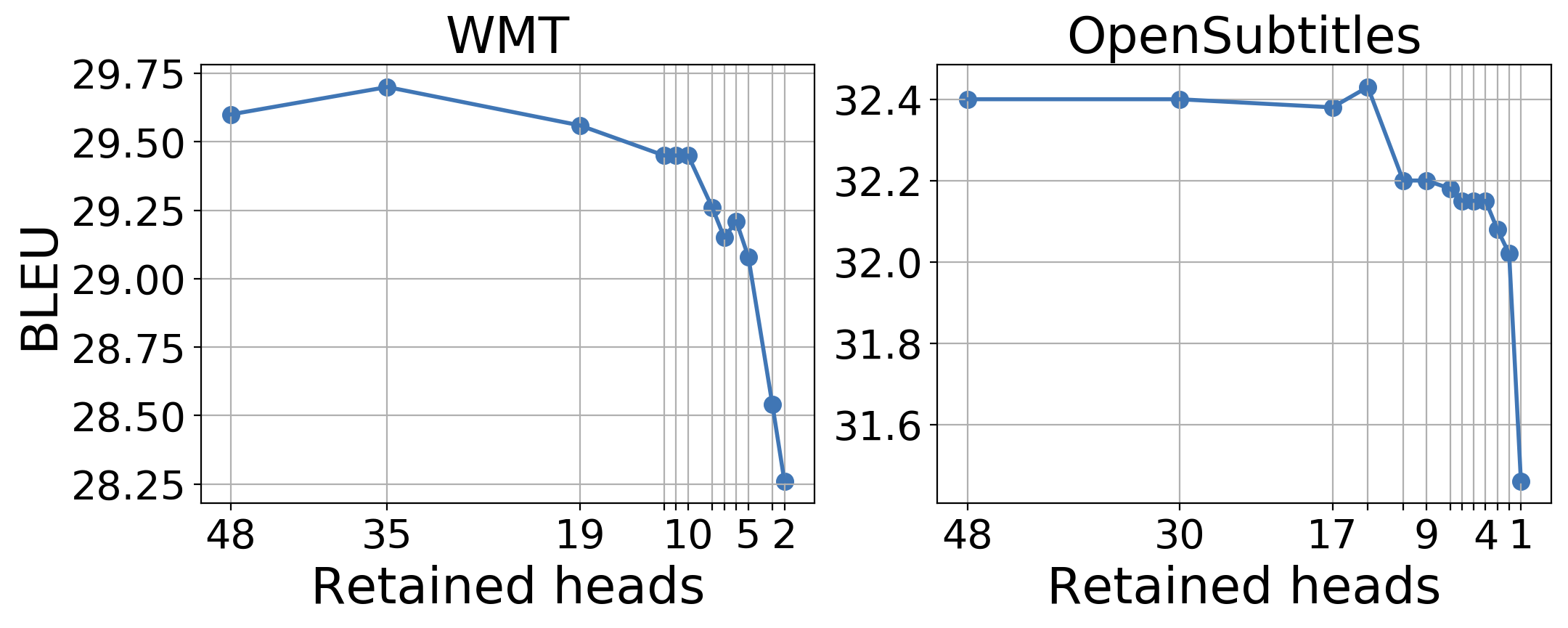}}
\caption{BLEU score as a function of number of retained encoder heads (EN-RU). Regularization applied by fine-tuning trained model.}
\label{fig:enc_heads_dying_bleu}
\end{figure}

BLEU scores are provided in  Figure~\ref{fig:enc_heads_dying_bleu}. Surprisingly, for OpenSubtitles, we lose only $0{.}25$ BLEU when we prune all but 4 heads out of 48.\footnote{If all heads in a layer are pruned, the only remaining connection to the previous layer is the residual connection.} For the more complex WMT task, 10 heads in the encoder are sufficient to stay within $0{.}15$ BLEU of the full model.

\subsubsection{Functions of retained heads}

\begin{figure}[t!]
\center{\includegraphics[scale=0.17]{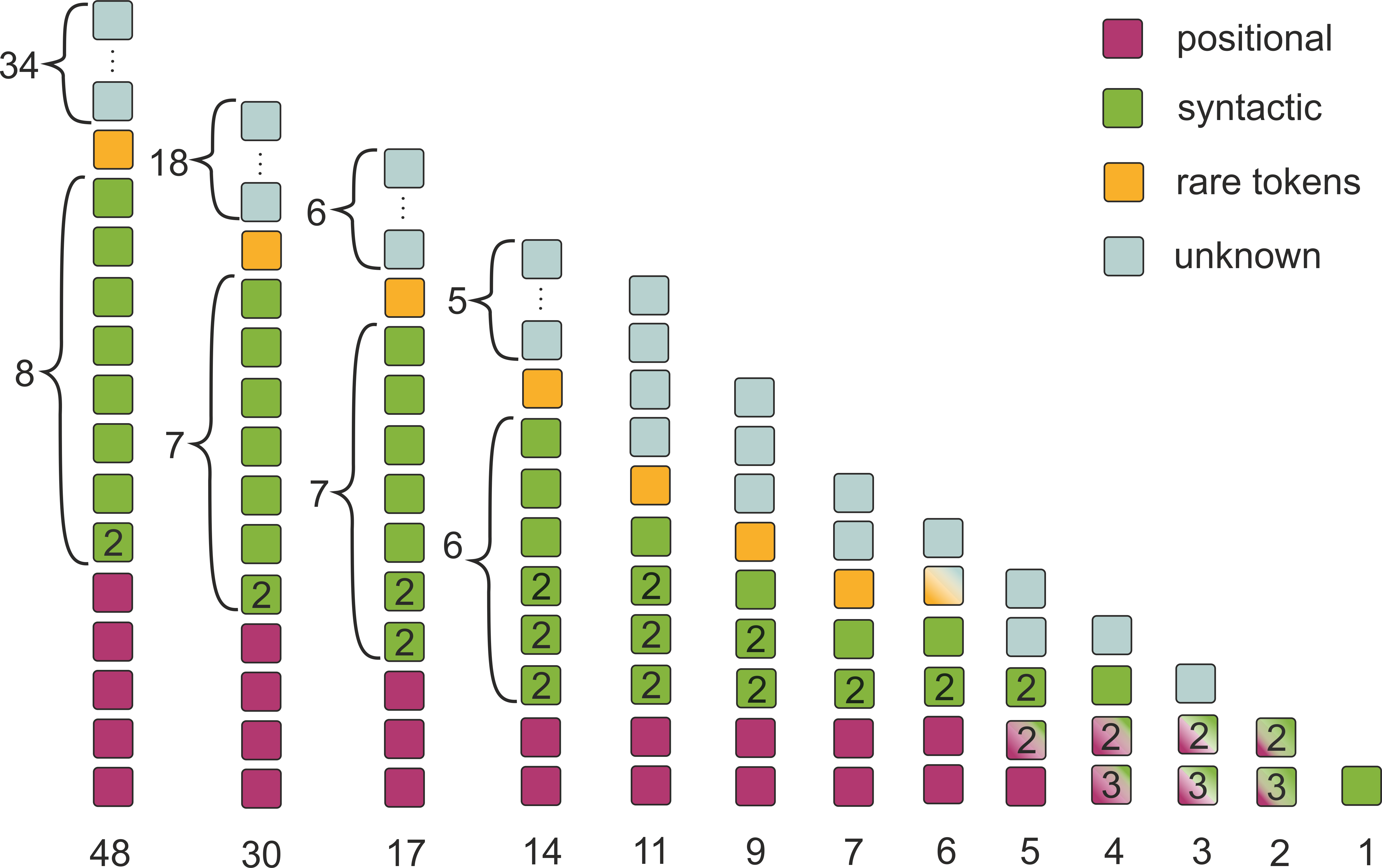}}
\caption{Functions of encoder heads retained after pruning. Each column represents all remaining heads after varying amount of pruning (EN-RU; Subtitles).}
%\vspace{-2ex}
\label{fig:encoder_heads_dying}
\end{figure}

Results in Figure~\ref{fig:enc_heads_dying_bleu} suggest that the encoder remains effective even with only 
a few heads. In this section, we investigate the function of those heads that remain in the encoder during pruning. 
Figure~\ref{fig:encoder_heads_dying} shows all heads color-coded for their function 
in a pruned model.
Each column corresponds to a model 
with a particular number of heads retained after pruning.
Heads from all layers are ordered by their function. Some heads can perform several functions (e.g., $s\rightarrow v$ and $v\rightarrow o$); in this case the number of functions is shown.

First, we note that the model with 17 heads retains heads with all the functions that we identified in Section~\ref{sect:characterizing_heads}, even though \sfrac{2}{3} of the heads have been pruned.

This indicates that these functions are indeed the most important. Furthermore, when we have fewer heads in the model, some functions ``drift'' to other heads: for example, we see positional heads starting to track syntactic dependencies; hence some heads are assigned more than one color at certain stages in Figure~\ref{fig:encoder_heads_dying}.

\subsection{Pruning all types of attention heads}
We found our pruning technique to be efficient at reducing the number of heads in the encoder without a major drop in translation quality. Now we investigate the effect of pruning all types of attention heads in the model (not just in the encoder). This allows us to evaluate the importance of different types of attention in the model for the task of translation.
In these experiments, we add gates to all multi-head attention heads in the Transformer, i.e.\ encoder and decoder self-attention and attention from the decoder to the encoder.

\subsubsection{Quantitative results: BLEU score}

\begin{table}[t!]
\centering
\begin{tabular}{lccc}
\toprule
 & \bf attention & \multicolumn{2}{c}{\bf BLEU} \\ 
 & \bf heads & from & from \\ 
 & (e/d/d-e) & trained & scratch \\ 
 \toprule
\multicolumn{3}{l}{\bf WMT, 2{.}5m}\\
\cmidrule{1-4}
baseline   &  48/48/48 & \multicolumn{2}{c}{\bf 29{.}6} \\
\cmidrule{1-4}
sparse heads   & 14/31/30 & 29{.}62 & 29{.}47 \\
   & 12/21/25 & 29{.}36 & 28{.}95 \\
   & 8/13/15 & 29{.}06 & 28{.}56 \\
   & 5/9/12 & 28{.}90 & 28{.}41 \\

\toprule
\multicolumn{3}{l}{\bf OpenSubtitles, 6m}\\
 \cmidrule{1-4}
baseline   &  48/48/48 & \multicolumn{2}{c}{\bf 32{.}4} \\
\cmidrule{1-4}
sparse heads   & 27/31/46 & 32{.}24 & 32{.}23 \\
   & 13/17/31 & 32{.}23 & 31{.}98 \\
   & 6/9/13 & 32{.}27 & 31{.}84 \\
\bottomrule
\end{tabular}\textbf{}
\caption{BLEU scores for gates in all attentions, EN-RU. Number of attention heads is provided in the following order: encoder self-attention, decoder self-attention, decoder-encoder attention.}
%\vspace{-2ex}
\label{tab:bleu_all_enru}
\end{table}

Results of experiments pruning heads in all attention layers are provided in Table~\ref{tab:bleu_all_enru}. For models trained on WMT data, we are able to prune almost \sfrac{3}{4} of encoder heads and more than \sfrac{1}{3} of heads in decoder self-attention and decoder-encoder attention without any noticeable loss in translation quality (sparse heads, row 1). We can also prune more than half of all heads in the model and lose no more than 0{.}25 BLEU.

While these results show clearly that the majority of attention heads can be removed from the fully trained model without significant loss in translation quality, it is not clear whether a model can be trained from scratch with such a small number of heads. In the rightmost column in Table~\ref{tab:bleu_all_enru} we provide BLEU scores for models trained with exactly the same number and configuration of heads in each layer as the corresponding pruned models but starting from a random initialization of parameters. 
Here the degradation in translation quality is more significant than for pruned models with the same number of heads. This agrees with the observations made in works on model compression: sparse architectures learned through pruning cannot be trained from scratch to the same test set performance as a model trained with joint sparsification and optimization~\cite{zhu2017prune,state-of-sparcity-2019}. In our case, attention heads are less likely to learn important roles when a model is retrained from scratch with a small number of heads.

\subsubsection{Heads importance}
\begin{figure}[t!]
\center{\includegraphics[scale=0.28]{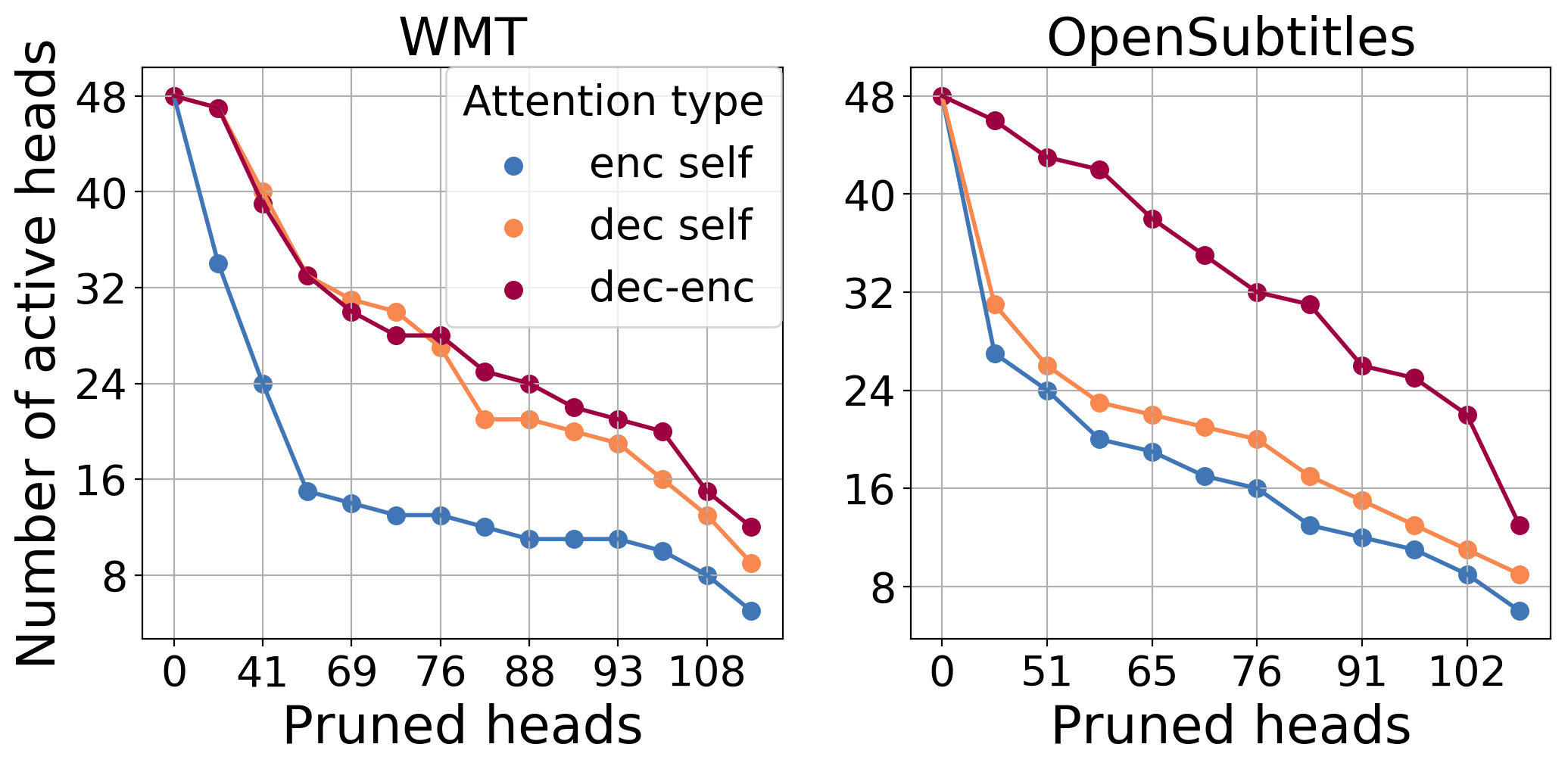}}
\caption{Number of active heads of different attention type for models with different sparsity rate}
\label{fig:all_heads_dying_by_attn_type}
\end{figure}

Figure~\ref{fig:all_heads_dying_by_attn_type} shows  the number
of retained heads for each attention type at different pruning rates.
We can see that the model prefers to prune encoder self-attention heads first, while decoder-encoder attention heads appear to be the most important for both datasets. Obviously, without decoder-encoder attention no translation can happen.

The importance of decoder self-attention heads, which function primarily as a target side language model, varies across domains. These heads appear to be almost as important as decoder-encoder attention heads for WMT data with its long sentences (24 tokens on average), and slightly more important than encoder self-attention heads for OpenSubtitles dataset where sentences are shorter (8 tokens on average).

\begin{figure}[t!]
\center{\includegraphics[scale=0.28]{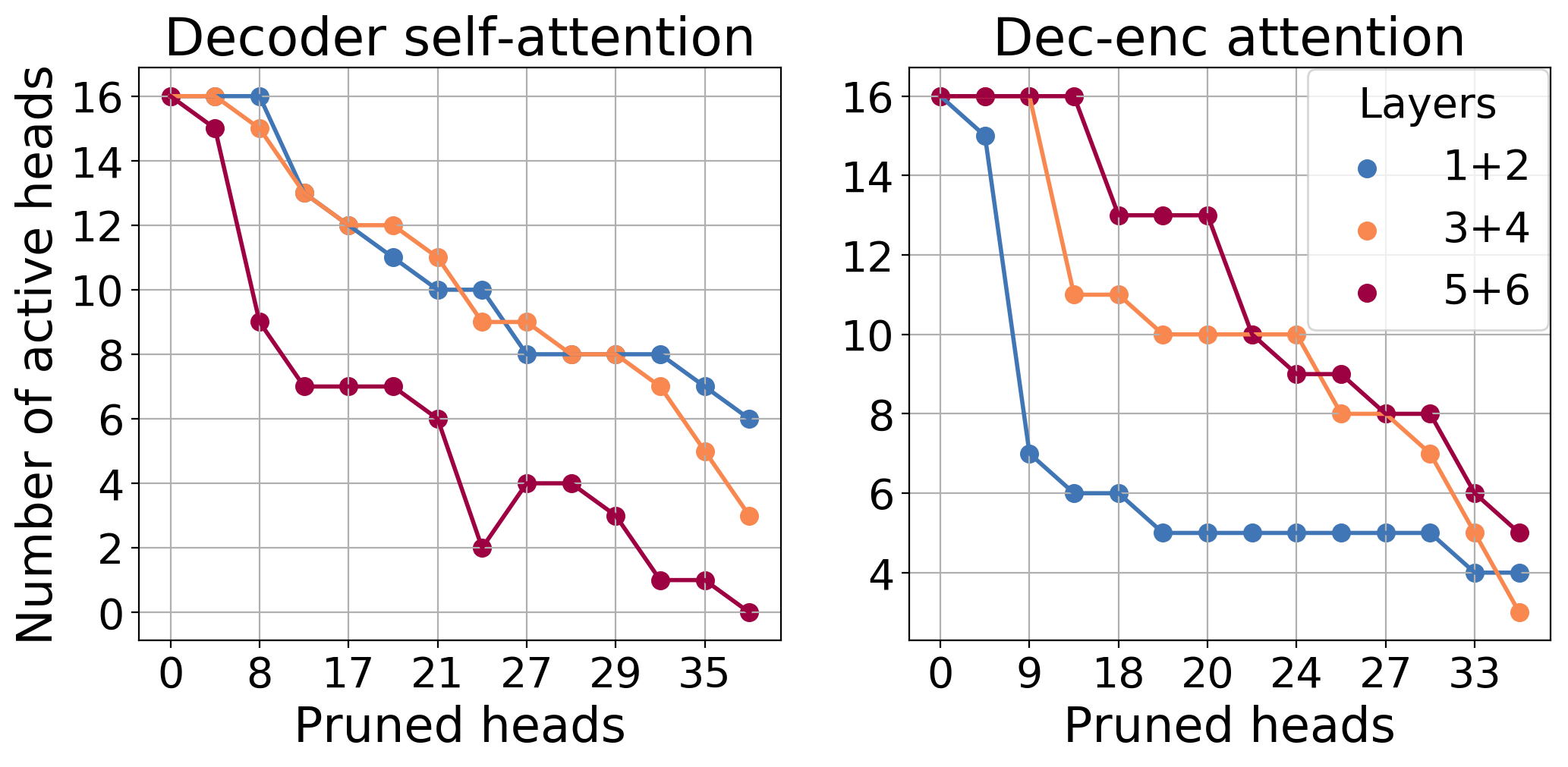}}
\caption{Number of active heads in different layers of the decoder for models with different sparsity rate (EN-RU, WMT)}
\label{fig:all_heads_dying_decoder}
\end{figure}

Figure~\ref{fig:all_heads_dying_decoder} shows the number of active self-attention and decoder-encoder attention heads at different layers in the decoder 
for models with different sparsity rate
(to reduce noise, we plot the sum of heads remaining in pairs of adjacent layers). It can be seen that self-attention heads are retained more readily in the lower layers, while decoder-encoder attention heads are retained in the higher layers. This suggests that lower layers of the Transformer's decoder are mostly responsible for language modeling, while higher layers are mostly responsible for conditioning on the source sentence. These observations are similar for both datasets we use.

\section{Related work}

One popular approach to
the analysis of NMT representations is to evaluate how informative they are for various linguistic tasks. Different levels of linguistic analysis have been considered including morphology~\cite{P17-1080,dalvi-morph-decoder,bisazza-tump:2018:EMNLP}, syntax~\cite{D16-1159} and semantics~\cite{hill-et-al,belinkov-I17-1001,raganato-tiedemann:2018:BlackboxNLP}.

\citet{bisazza-tump:2018:EMNLP} showed that the target language determines which information gets encoded. This agrees with our results for different domains on the English-Russian translation task in Section~\ref{sec:syntactic-heads:results}. There we observed that attention heads are more likely to track syntactic relations requiring more complex agreement in the target language (in this case the subject-verb relation).

An alternative method to study the ability of language models and machine translation models to capture hierarchical information is to test their sensitivity to specific grammatical errors  \cite{linzen-Q16-1037,colorless-green,tran-importance-of-being-recurrent,E17-2060,tang18-self-attention}.
While this line of work has shown that NMT models, including the Transformer, do learn some syntactic structures, our work provides further insight into the role of multi-head attention.

There are several works analyzing attention weights of different NMT models~\cite{monz-attention,voita18,tang-sennrich-nivre:2018:WMT,raganato-tiedemann:2018:BlackboxNLP}. \citet{raganato-tiedemann:2018:BlackboxNLP} use the self-attention weights of the Transformer's encoder to induce a tree structure for each sentence 
and compute the unlabeled attachment score of these trees. However they do not evaluate specific syntactic relations (i.e.\ labeled attachment scores) or consider how different heads specialize to specific dependency relations. 

Recently \citet{bau2019neurons-in-mt} proposed a method for identifying important individual neurons in NMT models. They show that similar important neurons emerge in different models. Rather than verifying the importance of individual neurons, we identify the importance of entire attention heads using layer-wise relevance propagation and verify our findings by observing which heads are retained when pruning the model.

\section{Conclusions}

We evaluate the contribution made by individual attention heads to Transformer model performance on translation. 
We use layer-wise relevance propagation to show that the relative contribution of heads varies: only a small subset of heads appear to be important for the translation task.
Important heads have one or more interpretable functions in the model, including attending to adjacent words and tracking specific syntactic relations.
To determine if the remaining less-interpretable heads are crucial to the model's performance, we introduce a new approach to pruning attention heads.

We observe that specialized heads are the last to be pruned, confirming their importance directly.
Moreover, the vast majority of heads, especially the encoder self-attention heads, can be removed without seriously affecting performance.
In future work, we would like to investigate how our pruning method compares to alternative methods of model compression in NMT.

\section*{Acknowledgments}
We would like to thank anonymous reviewers for their comments. We thank Wilker Aziz, Joost Bastings for their helpful suggestions. 
The authors also thank Yandex Machine Translation team for helpful discussions and inspiration. Ivan Titov acknowledges support of the European Research Council (ERC StG BroadSem 678254) and the Dutch National Science Foundation (NWO VIDI 639.022.518). 
%Rico Sennrich acknowledges support from the Swiss National Science Foundation (105212\_169888), the European Union’s Horizon 2020 research and innovation programme (grant agreement no 825460), and the Royal Society (NAF\textbackslash R1\textbackslash 180122). 

\nocite{training-tips-transformer}
\nocite{adam-optimizer}
\nocite{sennrich-bpe}

\bibliography{acl2019}

\begin{thebibliography}{34}
\expandafter\ifx\csname natexlab\endcsname\relax\def\natexlab#1{#1}\fi

\bibitem[{Bach et~al.(2015)Bach, Binder, Montavon, Klauschen, M{\"u}ller, and
  Samek}]{bach2015pixel}
Sebastian Bach, Alexander Binder, Gr{\'e}goire Montavon, Frederick Klauschen,
  Klaus-Robert M{\"u}ller, and Wojciech Samek. 2015.
\newblock On pixel-wise explanations for non-linear classifier decisions by
  layer-wise relevance propagation.
\newblock \emph{PloS one}, 10(7):e0130140.

\bibitem[{Bau et~al.(2019)Bau, Belinkov, Sajjad, Durrani, Dalvi, and
  Glass}]{bau2019neurons-in-mt}
Anthony Bau, Yonatan Belinkov, Hassan Sajjad, Nadir Durrani, Fahim Dalvi, and
  James Glass. 2019.
\newblock \href {https://openreview.net/pdf?id=H1z-PsR5KX} {Identifying and
  controlling important neurons in neural machine translation}.
\newblock In \emph{International Conference on Learning Representations}, New
  Orleans.

\bibitem[{Belinkov et~al.(2017{\natexlab{a}})Belinkov, Durrani, Dalvi, Sajjad,
  and Glass}]{P17-1080}
Yonatan Belinkov, Nadir Durrani, Fahim Dalvi, Hassan Sajjad, and James Glass.
  2017{\natexlab{a}}.
\newblock \href {https://doi.org/10.18653/v1/P17-1080} {What do neural machine
  translation models learn about morphology?}
\newblock In \emph{Proceedings of the 55th Annual Meeting of the Association
  for Computational Linguistics (Volume 1: Long Papers)}, pages 861--872.
  Association for Computational Linguistics.

\bibitem[{Belinkov et~al.(2017{\natexlab{b}})Belinkov, M{\`a}rquez, Sajjad,
  Durrani, Dalvi, and Glass}]{belinkov-I17-1001}
Yonatan Belinkov, Llu{\'i}s M{\`a}rquez, Hassan Sajjad, Nadir Durrani, Fahim
  Dalvi, and James Glass. 2017{\natexlab{b}}.
\newblock \href {http://aclweb.org/anthology/I17-1001} {Evaluating layers of
  representation in neural machine translation on part-of-speech and semantic
  tagging tasks}.
\newblock In \emph{Proceedings of the Eighth International Joint Conference on
  Natural Language Processing (Volume 1: Long Papers)}, pages 1--10. Asian
  Federation of Natural Language Processing.

\bibitem[{Bisazza and Tump(2018)}]{bisazza-tump:2018:EMNLP}
Arianna Bisazza and Clara Tump. 2018.
\newblock \href {http://www.aclweb.org/anthology/D18-1313} {The lazy encoder: A
  fine-grained analysis of the role of morphology in neural machine
  translation}.
\newblock In \emph{Proceedings of the 2018 Conference on Empirical Methods in
  Natural Language Processing}, pages 2871--2876, Brussels, Belgium.
  Association for Computational Linguistics.

\bibitem[{Bojar et~al.(2018)Bojar, Federmann, Fishel, Graham, Haddow, Huck,
  Koehn, and Monz}]{bojar-EtAl:2018:WMT1}
Ondřej Bojar, Christian Federmann, Mark Fishel, Yvette Graham, Barry Haddow,
  Matthias Huck, Philipp Koehn, and Christof Monz. 2018.
\newblock \href {http://www.aclweb.org/anthology/W18-6401} {Findings of the
  2018 conference on machine translation (wmt18)}.
\newblock In \emph{Proceedings of the Third Conference on Machine Translation,
  Volume 2: Shared Task Papers}, pages 272--307, Belgium, Brussels. Association
  for Computational Linguistics.

\bibitem[{Dalvi et~al.(2017)Dalvi, Durrani, Sajjad, Belinkov, and
  Vogel}]{dalvi-morph-decoder}
Fahim Dalvi, Nadir Durrani, Hassan Sajjad, Yonatan Belinkov, and Stephan Vogel.
  2017.
\newblock \href {http://aclweb.org/anthology/I17-1015} {Understanding and
  improving morphological learning in the neural machine translation decoder}.
\newblock In \emph{Proceedings of the Eighth International Joint Conference on
  Natural Language Processing (Volume 1: Long Papers)}, pages 142--151. Asian
  Federation of Natural Language Processing.

\bibitem[{{Ding} et~al.(2017){Ding}, {Liu}, {Luan}, and {Sun}}]{lrp-ding-2017}
Yanzhuo {Ding}, Yang {Liu}, Huanbo {Luan}, and Maosong {Sun}. 2017.
\newblock \href {https://doi.org/10.18653/v1/P17-1106} {Visualizing and
  understanding neural machine translation}.
\newblock In \emph{Proceedings of the 55th Annual Meeting of the Association
  for Computational Linguistics (Volume 1: Long Papers)}, pages 1150--1159,
  Vancouver, Canada. Association for Computational Linguistics.

\bibitem[{Gale et~al.(2019)Gale, Elsen, and Hooker}]{state-of-sparcity-2019}
Trevor Gale, Erich Elsen, and Sara Hooker. 2019.
\newblock \href {https://arxiv.org/abs/1902.09574} {The state of sparsity in
  deep neural networks}.
\newblock \emph{arXiv preprint}.

\bibitem[{Ghader and Monz(2017)}]{monz-attention}
Hamidreza Ghader and Christof Monz. 2017.
\newblock \href {http://aclweb.org/anthology/I17-1004} {What does attention in
  neural machine translation pay attention to?}
\newblock In \emph{Proceedings of the Eighth International Joint Conference on
  Natural Language Processing (Volume 1: Long Papers)}, pages 30--39. Asian
  Federation of Natural Language Processing.

\bibitem[{Gulordava et~al.(2018)Gulordava, Bojanowski, Grave, Linzen, and
  Baroni}]{colorless-green}
Kristina Gulordava, Piotr Bojanowski, Edouard Grave, Tal Linzen, and Marco
  Baroni. 2018.
\newblock \href {https://doi.org/10.18653/v1/N18-1108} {Colorless green
  recurrent networks dream hierarchically}.
\newblock In \emph{Proceedings of the 2018 Conference of the North American
  Chapter of the Association for Computational Linguistics: Human Language
  Technologies, Volume 1 (Long Papers)}, pages 1195--1205. Association for
  Computational Linguistics.

\bibitem[{Hill et~al.(2017)Hill, Cho, Jean, and Bengio}]{hill-et-al}
Felix Hill, Kyunghyun Cho, Sébastien Jean, and Y~Bengio. 2017.
\newblock \href {https://doi.org/10.1007/s10590-017-9194-2} {The
  representational geometry of word meanings acquired by neural machine
  translation models}.
\newblock \emph{Machine Translation}, 31.

\bibitem[{Jang et~al.(2017)Jang, Gu, and Poole}]{jang2017gumbel}
Eric Jang, Shixiang Gu, and Ben Poole. 2017.
\newblock \href {https://openreview.net/forum?id=rkE3y85ee} {Categorical
  reparameterization with gumbel-softmax}.
\newblock In \emph{International Conference on Learning Representations},
  Toulon, France.

\bibitem[{Kaiser and Bengio(2018)}]{kaiser2018discrete}
{\L}ukasz Kaiser and Samy Bengio. 2018.
\newblock Discrete autoencoders for sequence models.
\newblock \emph{arXiv preprint arXiv:1801.09797}.

\bibitem[{Kingma and Ba(2015)}]{adam-optimizer}
Diederik Kingma and Jimmy Ba. 2015.
\newblock {Adam: A method for stochastic optimization}.
\newblock In \emph{{Proceedings of the {International} {Conference} on
  {Learning} {Representation} (ICLR 2015)}}.

\bibitem[{Kingma and Welling(2014)}]{kingma-welling-vae}
Diederik~P. Kingma and Max Welling. 2014.
\newblock \href {https://openreview.net/forum?id=33X9fd2-9FyZd} {Auto-encoding
  variational bayes}.
\newblock In \emph{International Conference on Learning Representations},
  Banff, Canada.

\bibitem[{Linzen et~al.(2016)Linzen, Dupoux, and Goldberg}]{linzen-Q16-1037}
Tal Linzen, Emmanuel Dupoux, and Yoav Goldberg. 2016.
\newblock \href {http://aclweb.org/anthology/Q16-1037} {Assessing the ability
  of lstms to learn syntax-sensitive dependencies}.
\newblock \emph{Transactions of the Association for Computational Linguistics},
  4:521--535.

\bibitem[{Lison et~al.(2018)Lison, Tiedemann, and Kouylekov}]{LISON18.294}
Pierre Lison, Jörg Tiedemann, and Milen Kouylekov. 2018.
\newblock {OpenSubtitles2018: Statistical Rescoring of Sentence Alignments in
  Large, Noisy Parallel Corpora}.
\newblock In \emph{Proceedings of the Eleventh International Conference on
  Language Resources and Evaluation (LREC 2018)}, Miyazaki, Japan. European
  Language Resources Association (ELRA).

\bibitem[{Louizos et~al.(2018)Louizos, Welling, and
  Kingma}]{louizos2018learning}
Christos Louizos, Max Welling, and Diederik~P. Kingma. 2018.
\newblock \href {https://openreview.net/forum?id=H1Y8hhg0b} {Learning sparse
  neural networks through l\_0 regularization}.
\newblock In \emph{International Conference on Learning Representations},
  Vancouver, Canada.

\bibitem[{Maddison et~al.(2017)Maddison, Mnih, and Teh}]{maddison2017concrete}
Chris~J. Maddison, Andriy Mnih, and Yee~Whye Teh. 2017.
\newblock \href {https://openreview.net/forum?id=S1jE5L5gl} {The concrete
  distribution: A continuous relaxation of discrete random variables}.
\newblock In \emph{International Conference on Learning Representations},
  Toulon, France.

\bibitem[{Manning et~al.(2014)Manning, Surdeanu, Bauer, Finkel, Bethard, and
  McClosky}]{manning-EtAl:2014:P14-5}
Christopher~D. Manning, Mihai Surdeanu, John Bauer, Jenny Finkel, Steven~J.
  Bethard, and David McClosky. 2014.
\newblock \href {https://doi.org/10.3115/v1/P14-5010} {The {Stanford} {CoreNLP}
  natural language processing toolkit}.
\newblock In \emph{Proceedings of 52nd Annual Meeting of the Association for
  Computational Linguistics: System Demonstrations}, pages 55--60, Baltimore,
  Maryland. Association for Computational Linguistics.

\bibitem[{Niehues et~al.(2018)Niehues, Cattoni, Stüker, Cettolo, Turchi, and
  Federico}]{iwslt18-overview}
Jan Niehues, Ronaldo Cattoni, Sebastian Stüker, Mauro Cettolo, Marco Turchi,
  and Marcello Federico. 2018.
\newblock {The IWSLT 2018 Evaluation Campaign}.
\newblock In \emph{{Proceedings of the 15th International Workshop on Spoken
  Language Translation}}, pages 118--123, Bruges, Belgium.

\bibitem[{Popel and Bojar(2018)}]{training-tips-transformer}
Martin Popel and Ondrej Bojar. 2018.
\newblock \href {https://doi.org/10.2478/pralin-2018-0002} {{Training Tips for
  the Transformer Model}}.
\newblock pages 43--70.

\bibitem[{Raganato and Tiedemann(2018)}]{raganato-tiedemann:2018:BlackboxNLP}
Alessandro Raganato and Jörg Tiedemann. 2018.
\newblock \href {http://www.aclweb.org/anthology/W18-5431} {An analysis of
  encoder representations in transformer-based machine translation}.
\newblock In \emph{Proceedings of the 2018 EMNLP Workshop BlackboxNLP:
  Analyzing and Interpreting Neural Networks for NLP}, pages 287--297,
  Brussels, Belgium. Association for Computational Linguistics.

\bibitem[{Rezende et~al.(2014)Rezende, Mohamed, and
  Wierstra}]{pmlr-v32-rezende14}
Danilo~Jimenez Rezende, Shakir Mohamed, and Daan Wierstra. 2014.
\newblock \href {http://proceedings.mlr.press/v32/rezende14.html} {Stochastic
  backpropagation and approximate inference in deep generative models}.
\newblock In \emph{Proceedings of the 31st International Conference on Machine
  Learning}, volume~32 of \emph{Proceedings of Machine Learning Research},
  pages 1278--1286, Bejing, China. PMLR.

\bibitem[{Sennrich(2017)}]{E17-2060}
Rico Sennrich. 2017.
\newblock \href {http://aclweb.org/anthology/E17-2060.pdf} {{How Grammatical is
  Character-level Neural Machine Translation? Assessing MT Quality with
  Contrastive Translation Pairs}}.
\newblock In \emph{{Proceedings of the 15th Conference of the European Chapter
  of the Association for Computational Linguistics: Volume 2, Short Papers}},
  pages 376--382, Valencia, Spain.

\bibitem[{Sennrich et~al.(2016)Sennrich, Haddow, and Birch}]{sennrich-bpe}
Rico Sennrich, Barry Haddow, and Alexandra Birch. 2016.
\newblock \href {https://doi.org/10.18653/v1/P16-1162} {Neural machine
  translation of rare words with subword units}.
\newblock In \emph{{Proceedings of the 54th Annual Meeting of the Association
  for Computational Linguistics (Volume 1: Long Papers)}}, pages 1715--1725,
  Berlin, Germany. Association for Computational Linguistics.

\bibitem[{Shi et~al.(2016)Shi, Padhi, and Knight}]{D16-1159}
Xing Shi, Inkit Padhi, and Kevin Knight. 2016.
\newblock \href {https://doi.org/10.18653/v1/D16-1159} {Does string-based
  neural mt learn source syntax?}
\newblock In \emph{Proceedings of the 2016 Conference on Empirical Methods in
  Natural Language Processing}, pages 1526--1534. Association for Computational
  Linguistics.

\bibitem[{Tang et~al.(2018)Tang, M{\"u}ller, Rios, and
  Sennrich}]{tang18-self-attention}
Gongbo Tang, Mathias M{\"u}ller, Annette Rios, and Rico Sennrich. 2018.
\newblock \href {http://aclweb.org/anthology/D18-1458} {Why self-attention? a
  targeted evaluation of neural machine translation architectures}.
\newblock In \emph{Proceedings of the 2018 Conference on Empirical Methods in
  Natural Language Processing}, pages 4263--4272. Association for Computational
  Linguistics.

\bibitem[{{Tang} et~al.(2018){Tang}, {Sennrich}, and
  {Nivre}}]{tang-sennrich-nivre:2018:WMT}
Gongbo {Tang}, Rico {Sennrich}, and Joakim {Nivre}. 2018.
\newblock \href {http://aclweb.org/anthology/W18-6304} {An analysis of
  attention mechanisms: The case of word sense disambiguation in neural machine
  translation}.
\newblock In \emph{{Proceedings of the Third Conference on Machine Translation:
  Research Papers}}, pages 26--35, Belgium, Brussels. Association for
  Computational Linguistics.

\bibitem[{Tran et~al.(2018)Tran, Bisazza, and
  Monz}]{tran-importance-of-being-recurrent}
Ke~Tran, Arianna Bisazza, and Christof Monz. 2018.
\newblock \href {http://aclweb.org/anthology/D18-1503} {The importance of being
  recurrent for modeling hierarchical structure}.
\newblock In \emph{Proceedings of the 2018 Conference on Empirical Methods in
  Natural Language Processing}, pages 4731--4736. Association for Computational
  Linguistics.

\bibitem[{Vaswani et~al.(2017)Vaswani, Shazeer, Parmar, Uszkoreit, Jones,
  Gomez, Kaiser, and Polosukhin}]{attention-is-all-you-need}
Ashish Vaswani, Noam Shazeer, Niki Parmar, Jakob Uszkoreit, Llion Jones,
  Aidan~N Gomez, Lukasz Kaiser, and Illia Polosukhin. 2017.
\newblock \href
  {http://papers.nips.cc/paper/7181-attention-is-all-you-need.pdf} {Attention
  is all you need}.
\newblock In \emph{NeurIPS}, Los Angeles.

\bibitem[{{Voita} et~al.(2018){Voita}, {Serdyukov}, {Sennrich}, and
  {Titov}}]{voita18}
Elena {Voita}, Pavel {Serdyukov}, Rico {Sennrich}, and Ivan {Titov}. 2018.
\newblock \href {http://aclweb.org/anthology/P18-1117} {Context-aware neural
  machine translation learns anaphora resolution}.
\newblock In \emph{{Proceedings of the 56th Annual Meeting of the Association
  for Computational Linguistics (Volume 1: Long Papers)}}, pages 1264--1274,
  Melbourne, Australia. Association for Computational Linguistics.

\bibitem[{Zhu and Gupta(2017)}]{zhu2017prune}
Michael Zhu and Suyog Gupta. 2017.
\newblock \href {https://arxiv.org/abs/1710.01878} {To prune, or not to prune:
  exploring the efficacy of pruning for model compression}.
\newblock \emph{arXiv preprint arXiv:1710.01878}.

\end{thebibliography}
\bibliographystyle{acl_natbib}

%\newpage
\appendix

\section{Layer-wise Relevance Propagation}
\label{app:lrp}

Layer-wise relevance propagation (LRP) was originally designed to compute the contributions of single pixels to predictions of image classifiers~\cite{bach2015pixel}. LRP back-propagates relevance recursively from the output layer to the input layer.  We adapt LRP to the Transformer model to calculate relevance that measures the association degree between two arbitrary neurons in neural networks. In the following, we describe the general idea of the LRP method, give the formal definition used in our experiments and describe how to compute a head relevance.

\subsection{General idea}

Layer-wise relevance propagation in its general form assumes that the model can be decomposed into several layers of computation. 
The first layer are the inputs (for example, the pixels of an image or tokens of a sentence), the last layer is the real-valued prediction output of the model $f$. The $l$-th layer is modeled as a vector $z=(z_{d}^{(l)})_{d=1}^{V(l)}$ with dimensionality $V(l)$. Layer-wise relevance propagation assumes that we have a Relevance score $R_d^{(l+1)}$ for each dimension $z_{d}^{(l+1)}$ of the vector $z$ at layer $l + 1$. The idea is to find a Relevance score  $R_d^{(l)}$ for each dimension  $z_d^{(l)}$ of the vector $z$ at the next layer $l$ which is closer to the input layer such that the following equation holds:
$$f\!=\!\dots\!=\!\sum\limits_{d\in l + 1}\!R_d^{(l+1)}\!=\sum\limits_{d\in l}\!R_d^{(l)}=\dots=\sum\limits_{d}R_d^{(1)}.$$

This equation represents a \textit{conservation principle}, on which LRP relies to propagate the prediction back without using gradients. Intuitively, this means that total contribution of neurons at each layer is constant. Since we are interested only in heads relevance, we do not propagate till input variables and stop at the neurons of the encoder layer of interest.

\subsection{Formal rules}
In this section, we provide formal rules for propagating relevance. Here we follow the approach by~\citet{lrp-ding-2017} previously used for neural machine translation. 

Let $r_{u\leftarrow v}$ denote \textit{relevance} of neuron $u$ for neuron $v$.

\textbf{Definition 1} Given a neuron $u$, its incoming neuron set $IN(u)$ comprises all its direct connected preceding neurons in the network.

\textbf{Definition 2} Given a neuron $u$, its outcoming neuron set $OUT(u)$ comprises all its direct connected descendant neurons in the network.

\textbf{Definition 3} Given a neuron $v$ and its incoming neurons $u \in IN(v)$,  the \textit{weight ratio} measures the contribution of $u$ to $v$. It is calculated as
$$w_{u\rightarrow v}=\frac{W_{u,v}u}{\sum\limits_{u' \in IN(v)}\!\!\!\!W_{u',v}u'} \ \ \ \ \text{if} \ \  v=\!\!\!\!\sum\limits_{u' \in IN(v)}\!\!\!\!W_{u',v}u',$$
$$w_{u\rightarrow v}=\frac{u}{\sum\limits_{u' \in IN(v)} u'} \ \ \ \ \text{if} \ \  v=\!\!\!\!\prod\limits_{u' \in IN(v)} u'.$$
%$$ w_{u\rightarrow v} = 
%  \begin{cases} 
%   1 \!\!\!& \text{if } u=\!\!\!\max\limits_{u' \in IN(v)}\!\!u' \\
%   0 \!\!\!& \text{otherwise }
%  \end{cases} \ \ \text{if} \ \ v=\!\!\!\max\limits_{u' \in IN(v)}\!\!u'.
%$$
These equations define weight ratio for matrix multiplication and element-wise multiplication operations.

\textbf{Redistribution rule for LRP} Relevance is propagated using the \textit{local redistribution rule} as follows:
$$r_{u\leftarrow v} = \sum\limits_{z\in OUT(u)}w_{u\rightarrow z}r_{z\leftarrow v}.$$

The provided equations for computing weights ratio and the redistribution rule allow to compute the relative contribution of neurons at one point in a network to neurons at another. Note that we follow~\citet{lrp-ding-2017} and ignore non-linear activation functions. %because~\citet{bach2015pixel} indicate that LRP is invariant against the choice of non-linear function.

\subsection{Head relevance}
In our experiments, we compute relative contribution of each head to the network predictions. For this, we evaluate contribution of neurons in $\textnormal{head}_i$~(see equation~\ref{eq:mult_attention}) to the top-1 logit predicted by the model. Head relevance for a given prediction is computed as the sum of relevances of its neurons, normalized over heads in a layer. The final relevance of a head is its average relevance, where average is taken over all generation steps for a development set.

\section{Experimental setup}
\label{app:exp}

\subsection{Data preprocessing}

Sentences were encoded using byte-pair encoding~\cite{sennrich-bpe}, with source and target vocabularies of about 32000 tokens. For OpenSubtitles data, we pick only sentence pairs with a relative time overlap of subtitle frames between source and target language subtitles of at least $0.9$ to reduce noise in the data. 
Translation pairs were batched together by approximate sequence length. Each training batch contained a set of translation pairs containing approximately 16000\footnote{This can be reached by using several of GPUs or by accumulating the gradients for several batches and then making an update.} source tokens. It has been shown that Transformer's performance depends heavily on a batch size~\cite{training-tips-transformer}, and we chose a large value of batch size to ensure that models show their best performance.

\subsection{Model parameters}

We follow the setup of Transformer base model~\cite{attention-is-all-you-need}. More precisely, the number of layers in the encoder and in the decoder is $N=6$. We employ $h = 8$ parallel attention layers, or heads. The dimensionality of input and output is $d_{model} = 512$, and the inner-layer of a feed-forward networks has dimensionality $d_{ff}=2048$.

We use regularization as described in~\cite{attention-is-all-you-need}.

\subsection{Optimizer}
The optimizer we use is the same as in~\cite{attention-is-all-you-need}.
We use the Adam optimizer~\cite{adam-optimizer} with $\beta_1 = 0{.}9$, $\beta_2 = 0{.}98$ and $\varepsilon = 10^{-9}$. We vary the learning rate over the course of training, according to the formula:
\begin{multline*}
l_{rate}=scale\cdot \min(step\_num^{-0.5},\\ step\_num\cdot warmup\_steps^{-1.5}) 
\end{multline*}

We use $warmup\_steps = 16000$, $scale=4$.

\end{document}